\documentclass{article}

\usepackage{microtype}
\usepackage{graphicx}
\usepackage{booktabs} 

\usepackage{hyperref}

\usepackage[accepted]{icml2024}

\usepackage{amsmath}
\usepackage{amssymb}
\usepackage{mathtools}
\usepackage{amsthm}

\usepackage[capitalize,noabbrev]{cleveref}

\theoremstyle{plain}

\theoremstyle{definition}

\theoremstyle{remark}

\usepackage{amsmath,amsfonts,bm}

\def\eqref#1{equation~\ref{#1}}

\def\1{\bm{1}}

\def\rd{{\textnormal{d}}}

\def\rmJ{{\mathbf{J}}}

\def\rmS{{\mathbf{S}}}

\def\mI{{\bm{I}}}
\def\mJ{{\bm{J}}}

\DeclareMathAlphabet{\mathsfit}{\encodingdefault}{\sfdefault}{m}{sl}
\SetMathAlphabet{\mathsfit}{bold}{\encodingdefault}{\sfdefault}{bx}{n}

\newcommand{\R}{\mathbb{R}}

\newcommand{\Var}{\mathrm{Var}}

\DeclareMathOperator*{\rank}{rank}
\DeclareMathOperator*{\diag}{diag}

\usepackage{subcaption}
\usepackage{wrapfig}
\usepackage{multirow}
\usepackage{array}
\usepackage{tabularx}
\usepackage{float}
\usepackage{url}
\usepackage{xcolor}
\usepackage{fancyvrb}
\usepackage{makecell}

\usepackage{algorithm}
\usepackage{algpseudocode}
\algnotext{EndFunction}
\algnotext{EndIf}
\algnotext{EndFor}
\algnotext{EndWhile}

\newcommand{\leftcell}[1]{\makecell[tl]{#1}}

\newcommand{\vol}{\text{vol}}

\newcommand{\vect}[1]{\ensuremath{\mathbf{#1}}}

\newcommand{\manifold}{\mathcal{M}}

\newcommand{\LID}{\textrm{LID}}
\newcommand{\vecx}{\vect{x}}

\newcommand{\Mout}{\manifold_\text{out}}

\newcolumntype{Y}{>{\centering\arraybackslash}X}

\newcommand{\aref}[1]{\hyperref[#1]{Appendix~\ref*{#1}}}

\icmltitlerunning{A Geometric Explanation of the Likelihood OOD Detection Paradox}

\begin{document}

\twocolumn[
\icmltitle{A Geometric Explanation of the Likelihood OOD Detection Paradox}

\icmlsetsymbol{equal}{*}

\begin{icmlauthorlist}
\icmlauthor{Hamidreza Kamkari}{L6,UofT,Vector}
\icmlauthor{Brendan Leigh Ross}{L6}
\icmlauthor{Jesse C. Cresswell}{L6}
\icmlauthor{Anthony L. Caterini}{L6}
\icmlauthor{Rahul G. Krishnan}{UofT,Vector}
\icmlauthor{Gabriel Loaiza-Ganem}{L6}
\author{Hamidreza Kamkari$^{1, 2, 3}$, Brendan Leigh Ross$^{1}$, Jesse C. Cresswell$^{1}$,  Anthony L. Caterini$^{1}$, Rahul G. Krishnan$^{2, 3}$ \& Gabriel Loaiza-Ganem$^{1}$
}

\end{icmlauthorlist}

\icmlaffiliation{L6}{Layer 6 AI}
\icmlaffiliation{UofT}{University of Toronto}
\icmlaffiliation{Vector}{Vector Institute}

\icmlcorrespondingauthor{Hamidreza Kamkari, Brendan Leigh Ross, Jesse C. Cresswell, Anthony L. Caterini, Gabriel Loaiza-Ganem}{\{hamid, brendan, jesse, anthony, gabriel\}@layer6.ai}
\icmlcorrespondingauthor{Rahul G. Krishnan}{rahulgk@cs.toronto.edu}

\icmlkeywords{Out-of-distribution Detection, Deep Generative Models, Machine Learning}

\vskip 0.3in
]

\printAffiliationsAndNotice{} 

\begin{abstract}
{\looseness=-1}Likelihood-based deep generative models (DGMs) commonly exhibit a puzzling behaviour: 
when trained on a relatively complex dataset, they assign higher likelihood values to out-of-distribution (OOD) data from simpler sources. Adding to the mystery, OOD samples are never generated by these DGMs despite having higher likelihoods. This two-pronged paradox has yet to be conclusively explained, making likelihood-based OOD detection unreliable. Our primary observation is that high-likelihood regions will not be generated if they contain minimal probability mass. We demonstrate how this seeming contradiction of large densities yet low probability mass can occur around data confined to low-dimensional manifolds. We also show that this scenario can be identified through local intrinsic dimension (LID) estimation, and propose a method for OOD detection which pairs the likelihoods and LID estimates obtained from a \emph{pre-trained} DGM. 
Our method can be applied to normalizing flows and score-based diffusion models, and obtains results which match or surpass state-of-the-art OOD detection benchmarks using the same DGM backbones. 
Our code is available at \url{https://github.com/layer6ai-labs/dgm_ood_detection}.

\end{abstract}

\section{Introduction} \label{sec:introduction}
Out-of-distribution (OOD) detection \citep{quinonero2008dataset, rabanser2019failing, ginsberg2022learning} is crucial for ensuring the safety and reliability of machine learning models given their deep integration into real-world applications ranging from finance \citep{sirignano2019universal} to medical diagnostics \citep{esteva2017dermatologist}. In areas as critical as autonomous driving \citep{bojarski2016end} and medical imaging \citep{litjens2017survey, adnan2022federated}, these models, while proficient with in-distribution data, may give overconfident or plainly incorrect outputs when faced with OOD samples \citep{wei2022mitigating}. 

We focus on OOD detection using likelihood-based deep generative models (DGMs), which aim to learn the density that generated the observed data. Maximum-likelihood and related objectives operate by increasing model likelihoods or appropriate surrogates on training data, and since probability densities must be normalized, one might expect lower likelihoods for OOD points. Likelihood-based DGMs such as normalizing flows (NFs) \citep{dinh2016density, kingma2018glow, durkan2019neural} and diffusion models (DMs) \citep{sohl2015deep, ho2020denoising, song2020score, song2021maximum} have proven to be powerful DGMs that can render photo-realistic images. Given these successes, it seems reasonable to attempt OOD detection by thresholding the likelihood of a query datum under a trained model.

Surprisingly, likelihood-based DGMs trained on more complex datasets assign higher likelihoods to OOD datapoints taken from simpler datasets \citep{choi2018waic,nalisnick2018deep,havtorn2021hierarchical}. This becomes even more puzzling in light of the facts that $(i)$ said DGMs are trained to assign high likelihoods to in-distribution data without having been exposed to OOD data, and $(ii)$ they only generate samples which are visually much more similar to the training data. In this work, we explore the following explanation for how both these observations can simultaneously be true \citep{zhang2021understanding}:
\begin{center}
\emph{OOD datapoints can be assigned higher likelihoods while not being generated if they belong to regions of low probability mass.}
\end{center}

\begin{figure*}[th]\captionsetup{font=footnotesize}
    \centering
    \begin{minipage}[b]{\textwidth}
        \centering
        \begin{subfigure}[b]{0.33\textwidth}
        \includegraphics[width=1\textwidth, trim = 0 0 0 10, clip]{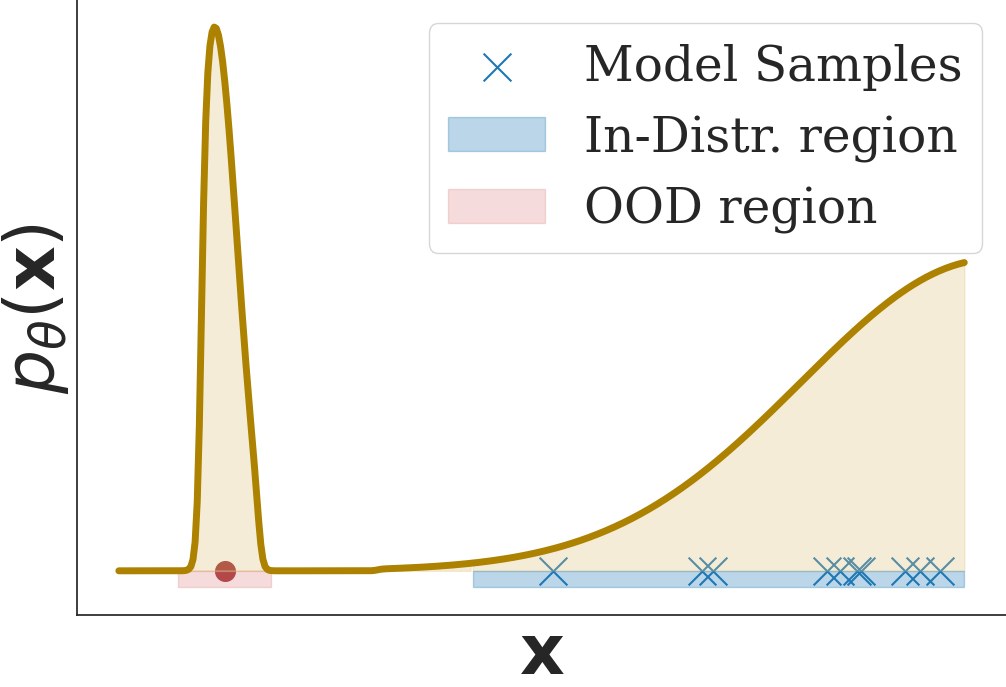}
        \caption{}
        \label{fig:main_figure_a}
        \end{subfigure}
        \hspace{50pt}
        \begin{subfigure}[b]{0.33\textwidth}
        \includegraphics[width=1\textwidth, trim = 0 40 0 65, clip]{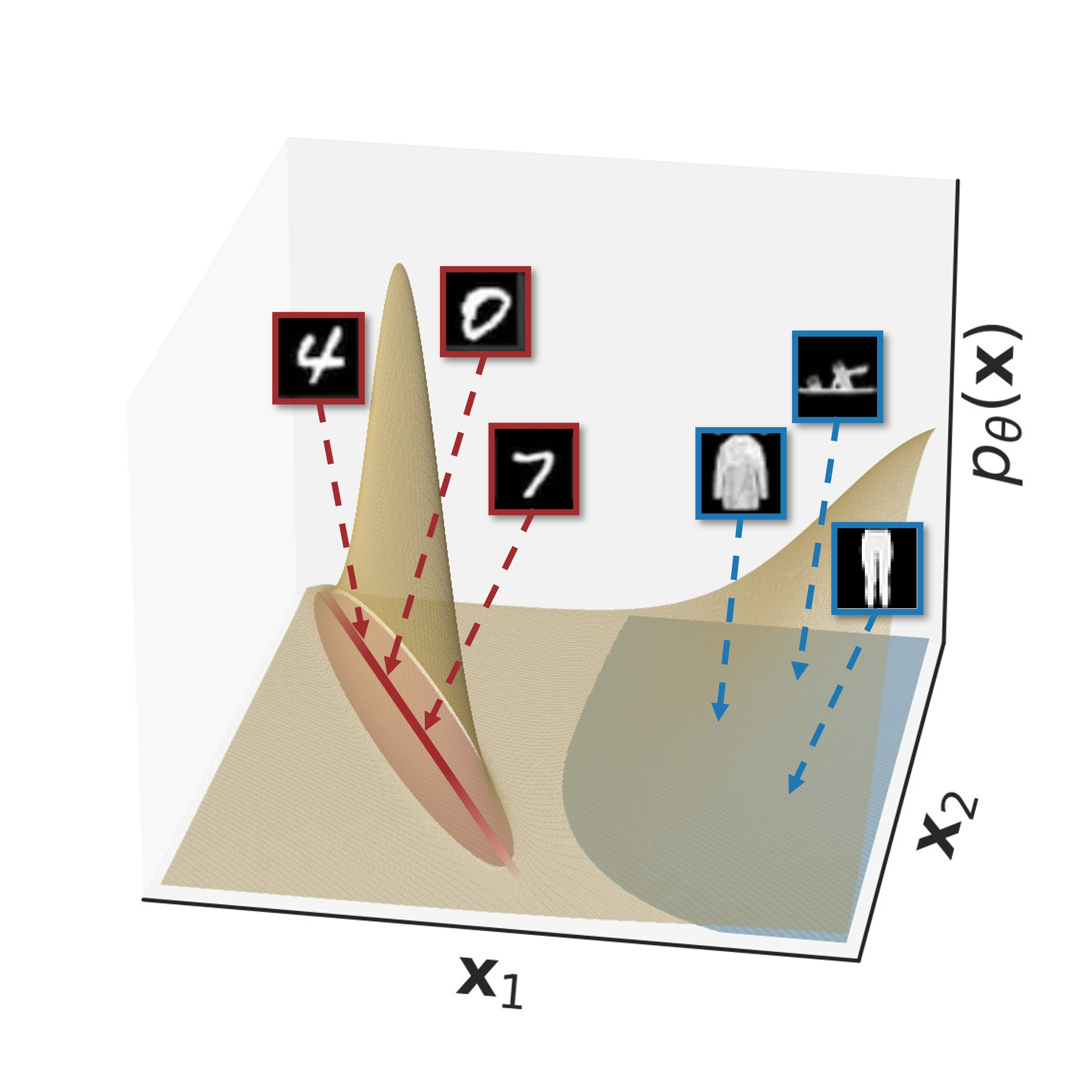}
        \caption{}
        \label{fig:main_figure_b}
        \end{subfigure}
    \end{minipage}
    \caption{
    \textbf{(a)} A 1D density which is highly peaked in the OOD region (red) assigns high likelihood, but low probability mass to OOD data.     \textbf{(b)} An analogous sketch for a 2D density concentrated around a 1D OOD manifold (red line), illustrated with FMNIST as in-distribution and MNIST as OOD. The model density has become sharply peaked around the manifold of ``simpler'' data which has low intrinsic dimension, which is nonetheless assigned lower probability mass as it has negligible volume.}
    \label{fig:main_figure}
    \vspace{5pt}
\end{figure*}

\autoref{fig:main_figure} illustrates that regions assigned high density by a model may integrate to very little probability mass. Our key insight is that when OOD data is ``simpler''  in the sense that it concentrates on a manifold of lower dimension than in-distribution data, the phenomenon depicted in \autoref{fig:main_figure} becomes completely consistent with empirical observations. Based on this insight, we develop a new understanding of the eponymous paradox, leading us to the realization that estimating the intrinsic dimension of the involved manifolds provides a simple and effective way to perform OOD detection \emph{using only a pre-trained likelihood-based DGM}.

\paragraph{Contributions} We $(i)$ develop an OOD detection method which classifies a datum as in-distribution only if it has high likelihood and is in a region with non-negligible probability mass -- as measured by a large local intrinsic dimension (LID) estimate of the DGM's learned manifold; $(ii)$ empirically verify our explanation for the OOD paradox for both NFs and DMs; $(iii)$ achieve or match state-of-the-art OOD detection performance among methods using the same DGM backbone as us.
\section{Background} \label{sec:background}

\paragraph{Normalizing Flows} In this study, we target DGMs that produce a density model $p_\theta$, parameterized by $\theta$, which can be easily evaluated. Among them, NFs readily provide probability densities through the change of variables formula, making them suitable for studying pathologies in the likelihood function. A NF is a diffeomorphic mapping $f_\theta: \mathcal{Z} \rightarrow \mathcal{X}$ from a latent space $\mathcal{Z}=\mathbb{R}^d$ to data space $\mathcal{X}=\mathbb{R}^d$, which transforms a simple distribution $p_Z$ on $\mathcal{Z}$, typically an isotropic Gaussian, into a complicated data distribution $p_\theta$ on $\mathcal{X}$. The change of variables formula allows one to evaluate the likelihood of a datum $\vect{x} \in \mathcal{X}$ as
\begin{equation} \label{eq:change-of-variables}
\log p_\theta(\vect{x}) = \log p_Z(\vect{z}) - \log \left| \det \mJ(\vect{z}) \right|,
\end{equation}
where $\vect{z} = f^{-1}_\theta(\vect{x})$ and $\mJ(\vect{z}) = \nabla_{\vect{z}} f_\theta (\vect{z}) \in \R^{d \times d}$. 
NFs are constructed such that $\det \mJ(\vect{z})$ can be efficiently evaluated, and are trained through maximum-likelihood, $\max_\theta \mathbb{E}_{\vect{x} \sim p_0}[\log p_\theta(\vect{x})]$, where $p_0$ is the true data-generating distribution. 
Sampling $\vect{x} \sim p_\theta$ is achieved by transforming a sample $\vect{z} \sim p_Z$ through $f_\theta$, i.e.\ $\vect{x}{=}f_\theta(\vect{z})$.

\begin{figure*}\captionsetup{font=footnotesize}
    \centering
    \begin{minipage}[b]{\textwidth}
        \centering
        \begin{subfigure}[b]{0.44\textwidth}
            \includegraphics[width=\textwidth]{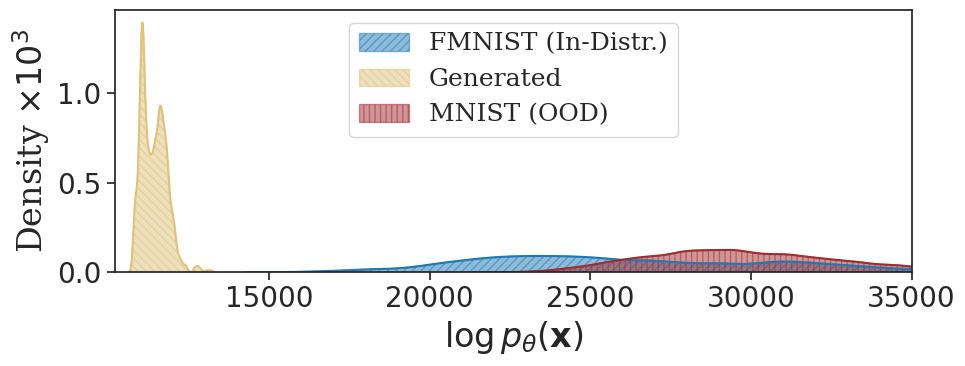}
            \caption{FMNIST-trained DM vs.\ MNIST.}
            \label{fig:pathology_fmnist_vs_mnist_omniglot_diffusion}
        \end{subfigure}
        \hspace{10pt}
        \begin{subfigure}[b]{0.44\textwidth}
            \includegraphics[width=\textwidth]{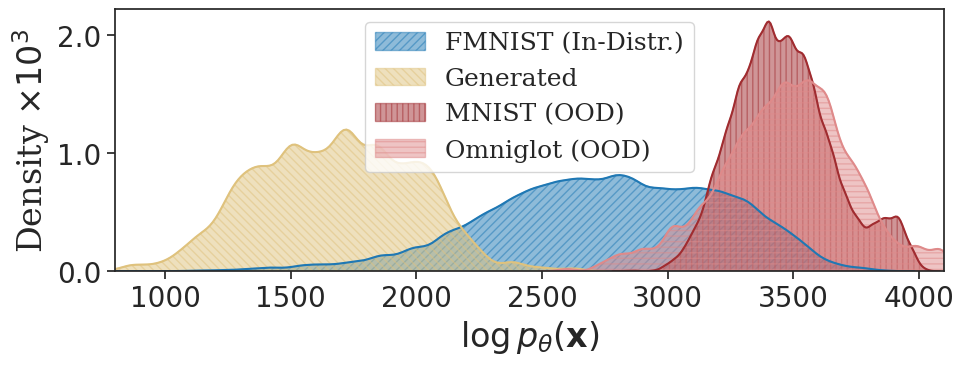}
            \caption{FMNIST-trained NF vs.\ MNIST and Omniglot.}
            \label{fig:pathology_fmnist_vs_mnist_omniglot_flow}
        \end{subfigure}
    \end{minipage}
    
    \vspace{1mm}
    \begin{minipage}[b]{\textwidth}
        \centering
        \begin{subfigure}[b]{0.29\textwidth}
            \includegraphics[width=\textwidth]{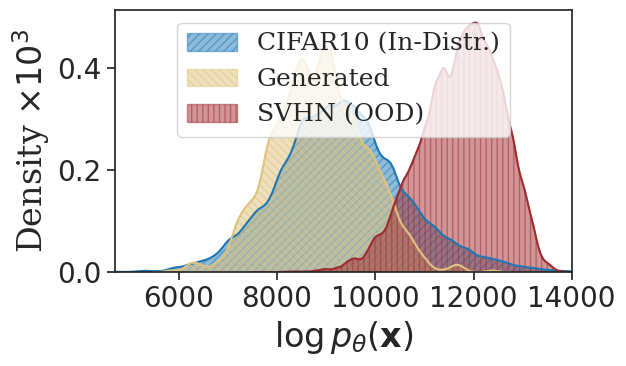}
            \caption{CIFAR10-trained NF vs.\ SVHN.}
            \label{fig:pathology_cifar10_vs_svhn_flow}
        \end{subfigure}
        \hspace{12pt}
        \begin{subfigure}[b]{0.29\textwidth}
            \includegraphics[width=\textwidth]{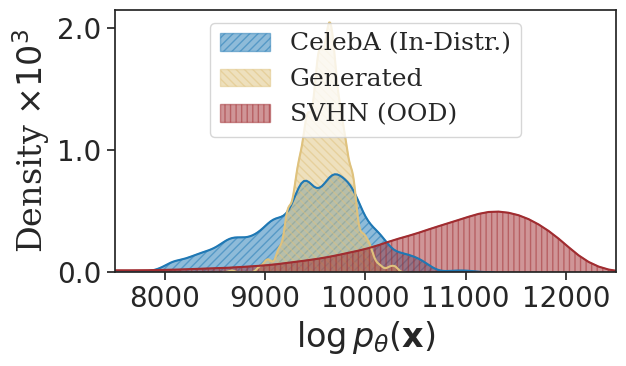}
            \caption{CelebA-trained NF vs.\ SVHN.}
            \label{fig:pathology_celeba_vs_svhn_flow}
        \end{subfigure}
        \hspace{12pt}
        \begin{subfigure}[b]{0.29\textwidth}
            \includegraphics[width=\textwidth]{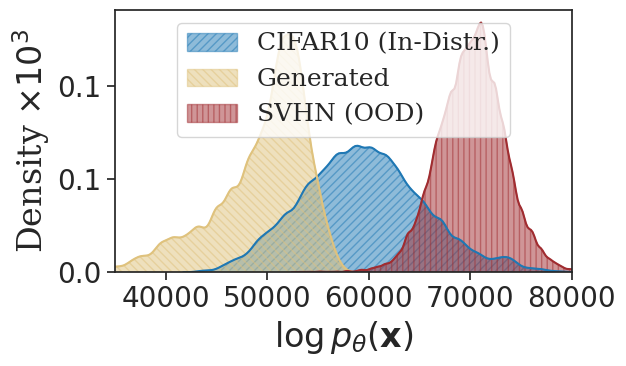}
            \caption{CIFAR10-trained DM vs.\ SVHN.}
            \label{fig:pathology_cifar10_vs_svhn_diffusion}
        \end{subfigure}
    \end{minipage}
    \caption{\textbf{(a)} A FMNIST-trained DM assigns higher likelihoods to MNIST. \textbf{(b)} A NF trained on FMNIST shows notably lower likelihoods on its own generated samples than on OOD data. \textbf{(c-e)} Analogous pathologies on RGB datasets, both for DMs and NFs.}
    \label{fig:likelihood_pathologies}
    \vspace{-5pt}
\end{figure*}

\paragraph{Diffusion Models} DMs are popular and also admit likelihood evaluation. Various formulations of DMs exist; here we use score-based models \citep{song2020score}. DMs first define an It\^o stochastic differential equation (SDE):
\begin{equation}\label{eq:forward_sde}
    \rd \vect{x}_t = h(\vect{x}_t, t) \rd t + g(t) \rd \vect{w}_t,\quad \vect{x}_0 \sim p_0,
\end{equation}
where $h: \mathcal{X} \times [0, T] \rightarrow \mathcal{X}$, $g:[0, T] \rightarrow \R$, and $T>0$ are hyperparameters, and where $\mathbf{w}_t$ denotes a $d$-dimensional Brownian motion. This SDE prescribes how to transform data $\vect{x}_0$ into noisy data $\vect{x}_t$, whose distribution we denote as $p_t$, the intuition being that $p_T$ is extremely close to ``pure noise''. \autoref{eq:forward_sde} can be reversed in time in the sense that $\vect{y}_t = \vect{x}_{T-t}$ obeys the SDE
\begin{equation}
\begin{aligned}\label{eq:backward_sde_exact}
    \rd \vect{y}_t = & \left(g(T-t)^2 s_{T-t}(\vect{y}_t) -h(\vect{y}_t, T-t)\right)\rd t \\
    & + g(T-t) \rd \bar{\vect{w}}_t, \quad \vect{y}_0 \sim p_T,
\end{aligned}
\end{equation}
where $s_{T-t}(\vect{y}_t)=\nabla_{\vect{y}_t} \log p_{T-t}(\vect{y}_t)$ denotes the (Stein) score, and $\bar{\vect{w}}_t$ is another Brownian motion. Solving \autoref{eq:backward_sde_exact} would result in samples from $p_0$ at time $T$, but both the score and $p_T$ are unknown. DMs use a neural network $s_\theta: \mathcal{X} \times [0, T] \rightarrow \mathcal{X}$ whose goal is to learn the true score. This is achieved through the denoising score matching objective \citep{vincent2011connection}, which \citet{song2021maximum} showed can be interpreted as likelihood-based for an appropriate hyperparameter choice. Sampling from a trained DM is achieved by approximately solving \autoref{eq:backward_sde_exact}: $s_\theta(\vect{y}_t, T-t)$ replaces $s_{T-t}(\vect{y}_t)$, and a Gaussian distribution $\hat{p}_T$ with an appropriately chosen covariance replaces $p_T$. This procedure implicitly defines the density $p_\theta$ of a DM. \citet{song2020score} show that DMs can be interpreted as continuous NFs \citep{chen2018neural}, transforming samples from $\hat{p}_T$ into (approximate) samples from $p_0$. In turn, this enables evaluating $p_\theta$ through a corresponding change of variable formula analogous to \autoref{eq:change-of-variables}.

\vspace{-5pt}
\paragraph{Likelihood Pathologies in OOD Detection} \citet{choi2018waic} and \citet{nalisnick2018deep} first uncovered unintuitive behaviour that pervasively affects likelihood-based DGMs. For instance, NFs trained on relatively complex datasets like CIFAR10 \citep{krizhevsky2009learning} and FMNIST \citep{xiao2017fashion} often yield high likelihoods when tested on simpler ones like SVHN \citep{netzer2011reading} and MNIST \citep{lecun2010mnist}, respectively, despite the latter datasets not having been seen in the training process. While this issue is not exclusive to images \citep{ren2019likelihood}, our experiments, shown in \autoref{fig:likelihood_pathologies}, confirm these previous findings for models trained on image data. Additionally, we find that this pathological behaviour is not limited to these well-known cases, but extends to numerous dataset pairs and generated samples (see \autoref{appx:extra-likelihood-pathologies} for details).

\vspace{-6pt}
\paragraph{Local Intrinsic Dimension} According to the manifold hypothesis, natural data lies around low-dimensional submanifolds of $\mathcal{X}=\mathbb{R}^d$ \citep{bengio2013representation, pope2021intrinsic}, where $d$ is the \textit{ambient} dimension of the data space. 
The \textit{local intrinsic dimension} (LID) of $\vect{x} \in \mathcal{X}$ with respect to these data submanifolds is the dimension of the submanifold that contains $\vect{x}$. For example, if the ambient space is $\R^2$ and the data manifold is the 1D unit circle $S^1$, then any point $\vect{x} \in S^1$ will have a local intrinsic dimension of $1$. Often, datasets concentrate on multiple non-overlapping submanifolds of different dimensionalities \citep{brown2022verifying}, in which case the LID will vary between datapoints.

Density models $p_\theta$ implicitly attempt to learn these manifolds by accumulating probability mass around them. As a consequence, even when defined on the full $d$-dimensional space $\mathcal{X}$, trained densities $p_\theta$ implicitly encode low-dimensional manifold structure. We refer to the manifold implied by $p_\theta$ as $\manifold_\theta$, which informally corresponds to regions of high density. When referring to LID with respect to the implied manifold $\manifold_\theta$, we will write $\LID_\theta(\vect{x})$; it will be of interest to estimate $\LID_\theta(\vect{x})$ for in- and out-of-distribution query points $\vect{x}$.

Sample-based methods to estimate intrinsic dimension exist \citep{fukunaga1971algorithm, levina2004maximum, johnsson2014low, facco2017estimating, bac2021}. Unfortunately, most of these are inadequate for our purposes, either because they estimate global (i.e.\ averaged or aggregated) intrinsic dimension instead of $\LID_\theta(\vect{x})$, or because they require observed samples around $\vect{x}$ to produce the estimate. Since we will want $\LID_\theta(\vect{x})$ for OOD points $\vect{x}$, the latter methods would require access to samples from $p_\theta$ which fall in the OOD region, which are of course unavailable. The key to circumvent this issue is to move away from sample-based estimators and instead rely directly on the given DGM.

\citet{tempczyk2022lidl} proposed such an estimator of LID. Unfortunately, it requires training multiple NFs, rendering it incompatible with our OOD detection goal of using a single pre-trained model. Meanwhile, \citet{stanczuk2022your} construct an estimator requiring a single \emph{variance exploding} DM, i.e.\ they set $h$ to zero in \autoref{eq:forward_sde}. They argue that, given a query $\vect{x}$, a small enough $t_0 > 0$, and $\vect{x}'$ sufficiently close to $\vect{x}$, $s_{\theta}(\vect{x}', t_0)$ will lie on the normal space (at $\vect{x}$) of the manifold containing $\vect{x}$ -- i.e.\ $s_{\theta}(\vect{x}', t_0)$ is orthogonal to the manifold. They propose using $k$ independent runs of \autoref{eq:forward_sde}, starting at $\vect{x}$ and evolving until time $t_0$, to obtain $\vect{x}^1_{t_0}, \dots, \vect{x}^k_{t_0}$ from which they construct the matrix $\rmS(\vect{x}) = [s_\theta(\vect{x}^1_{t_0}, t_0)|\cdots|s_\theta(\vect{x}^k_{t_0}, t_0)] \in \R^{d \times k}$. The rank of $\rmS(\vect{x})$ estimates the dimension of the normal space when $k$ is large enough. In turn, LID can be estimated as
\begin{equation}\label{eq:lid_stan}
    \LID_\theta(\vect{x}) \approx d - \rank \rmS(\vect{x}).
\end{equation}
 
\section{Method} \label{sec:method}

Intuitively, the fact that DGMs never generate OOD samples suggests that they contain the information needed to discern between OOD and in-distribution data -- \emph{even when they assign higher likelihoods to the former}. In this section we will show how to leverage LID to extract this information from a pre-trained model. Although conceptually our insights are agnostic to the type of model being used and thus apply to all likelihood-based DGMs, we will focus on NFs and DMs as LID and likelihoods can be readily computed for them.

\subsection{LID, Volume, and Probability Mass} \label{subsec:lid-volume}
Before showing that $\LID_\theta(\vect{x})$ is the key to the OOD detection paradox, we heuristically explain here how $\LID_\theta(\vect{x})$ is related to the contiguous volume associated to the region around $\vect{x}$ by $p_\theta$. 

To demonstrate, consider a fitted three-dimensional model $p_{\theta}(\vect{x}) = w_\text{in}\mathcal{N}(\vect{x}; \mu_\text{in}, \Sigma_\text{in}) + w_\text{out}\mathcal{N}(\vect{x}; \mu_\text{out}, \Sigma_\text{out})$, where $\mathcal{N}(\vect{x}; \mu, \Sigma)$ is a Gaussian density with mean $\mu$ and covariance $\Sigma$ evaluated at $\vect{x}$. Let $\Vert \mu_\text{in} - \mu_\text{out} \Vert_2 \gg 0$ so that the $\mu_\text{out}$ component is sufficiently far from $\mu_\text{in}$, and let $w_\text{out} = \delta$ with $w_\text{in} = 1 - \delta$, where $1 \gg \delta > 0$ is vanishingly small so that data around $\mu_\text{out}$ has near-zero probability of being sampled. In these ways, $\vect{\mu}_\text{in}$ and $\vect{\mu}_\text{out}$ are analogous to in-distribution and pathological OOD data. Let the respective covariance matrices of the components be $\Sigma_{\text{in}} = \diag(\sigma^2, \sigma^2, \varepsilon^2)$ and $\Sigma_{\text{out}} = \diag(\varepsilon^2, \sigma^2, \varepsilon^2)$, where $\sigma \gg \varepsilon > 0$, with $\varepsilon$ very close to zero. The impact of these covariance matrices is that the model distribution $p_{\theta}$ 
extends primarily in two dimensions around $\mu_\text{in}$ and in a single dimension around $\mu_\text{out}$. Thus, numerically, $\LID_{\theta}(\mu_\text{in}) = 2$ and  $\LID_{\theta}(\mu_\text{out}) = 1$.

We first note that $p_{\theta}(\mu_\text{out}) > p_{\theta}(\mu_\text{in})$ whenever $\varepsilon < \sigma\delta$, meaning it is possible here for $p_{\theta}$ to be pathologically high on OOD data. Since, informally, ``probability mass = density $\times$ volume'', a reasonable way to measure the volume assigned by $p_\theta$ around each of these modes is the ratio of probability mass to density. For $i \in \{\text{in}, \text{out}\}$, mode $i$ has a respective mass of approximately $w_i$, so we can write
\begin{equation}\label{eq:vol}
    \text{\small $\vol_{\theta}(\mu_i) \approx \frac{w_i}{p_{\theta}(\mu_i)} \approx \frac{1}{\mathcal{N}(\mu_i; \mu_i, \Sigma_i)} \propto \varepsilon^3\left(\frac{\sigma}{\varepsilon}\right)^{\LID_{\theta}(\mu_i)}$}
    ,
\end{equation}
where, since $\sigma > \varepsilon$, volume monotonically increases with LID. Though the models we use throughout this work are much more complex, we carry forward the same intuition that the volume $\vol_\theta(\vect{x})$ assigned to the vicinity of $\vect{x}$ by $p_\theta$ increases with $\LID_{\theta}(\vect{x})$. A small $\LID_{\theta}(\vect{x})$ makes it possible in practice for $p_\theta(\vecx)$ to be high, despite $p_\theta$ assigning negligible probability mass around $\vecx$.

\paragraph{LID and probability mass} We now make the connection between LID and probability mass more concrete. For $R > 0$, we denote the $d$-dimensional Euclidean ball of radius $R$ centred at $\vect{x}$ as $B_R(\vect{x})$.
In \autoref{appx:probability-mass-and-lid} we argue that, for sufficiently negative scalars $r$,
\begin{equation}\label{eq:lid-vs-prob-mass}
    \dfrac{\partial}{\partial r} \log \int_{B_{e^r \sqrt{d}}(\vect{x})} p_\theta(\vect{x}') \rd \vect{x}' \approx \LID_\theta(\vect{x}) + C,
\end{equation}
where $C$ is a constant that depends neither on $\theta$ nor on $\vect{x}$. The integral on the left hand side corresponds to the probability assigned by $p_\theta$ to a small ball around $\vect{x}$. Thus, a large $\LID_\theta(\vect{x})$ is equivalent to a rapid growth of the log probability mass that $p_\theta$ assigns to a neighbourhood of $\vect{x}$ as the size of the neighbourhood increases. In turn, we should expect the probability mass assigned around $\vect{x}$ to be large if and only if both $p_\theta(\vect{x})$ and the aforementioned rate of change are large as well, meaning that $\LID$ can indeed be informally understood as monotonically related to the probability mass. This view of LID provides the same intuition as the more informal ``volume''-based interpretation, namely that probability mass being large is equivalent to both density and LID being large.

\subsection{Detecting OOD Data with LID} \label{subsec:detecting-ood}

We now discuss the situation illustrated in \autoref{fig:main_figure}. We begin by highlighting that there are three manifolds (or rather unions thereof) at play -- $\mathcal{M}_{\text{in}}$, $\mathcal{M}_{\text{out}}$, and $\mathcal{M}_{\theta}$ -- around which $p_0$, OOD data, and $p_\theta$ concentrate, respectively. We take the viewpoint that $\mathcal{M}_{\text{in}}$ and $\mathcal{M}_{\text{out}}$ do not overlap (i.e., $\mathcal{M}_{\text{in}} \cap \mathcal{M}_{\text{out}} = \emptyset$), otherwise OOD detection would be ill-posed \citep{le2021perfect}. The paradoxical nature of likelihood-based OOD detection can be summarized as follows: 
when $\vect{x} \in \mathcal{M}_{\text{out}}$ we should expect the ground truth $p_0(\vect{x}) \approx 0$  because $\mathcal{M}_{\text{in}} \cap \mathcal{M}_{\text{out}} = \emptyset$, and since $p_\theta$ was trained to approximate $p_0$, we should also expect $p_\theta(\vect{x}) \approx 0$. However, it is often observed that $p_\theta$ is \emph{larger} on $\mathcal{M}_{\text{out}}$ than on $\mathcal{M}_{\text{in}}$; i.e., $\mathcal{M}_{\text{out}} \subset \manifold_\theta$. We now explain how this behaviour can occur by leveraging the notion of volume assigned by $p_\theta$.

By the manifold hypothesis, $\mathcal{M}_{\text{out}}$ and $\mathcal{M}_{\text{in}}$ are low-dimensional, and they thus have zero (Lebesgue) volume in ambient space. However, here we are concerned with a different notion of ``volume'': the contiguous ``volume'' associated to a region around $\vect{x}$ by the full-dimensional density $p_\theta$. We informally define this ``volume'' as a ratio of probability mass to density, as in \autoref{eq:vol}. In the OOD paradox, $p_\theta(\vecx)$ is large for $\vect{x} \in \Mout$, yet samples are never drawn from $\Mout$, suggesting negligible probability mass has been assigned around the region. As a consequence, $p_\theta$ must have assigned a very small ``volume'' to the region around $\vecx$. This is made mathematically possible by the fact that $\Mout$ has a (Lebesgue) volume of zero, and thus $p_\theta$ can assign arbitrarily small ``volume'' to the region around $\Mout$, even when high densities are present. From this logic, we see that \emph{the paradox is fully characterized by $p_\theta$ assigning high density to, but low ``volume'' around, the point $\vecx \in \Mout$.}

\emph{A priori}, $\mathcal{M}_{\text{out}}$ being contained in the region over which $p_\theta$ happens to behave pathologically (i.e., $\mathcal{M}_{\text{out}} \subset \mathcal{M}_\theta$)  might seem like an unbelievable coincidence. However, in the case of NFs, past work by \citet{kirichenko2020normalizing} and \citet{schirrmeister2020understanding} has shown that the multi-scale convolutional architecture used by these models fixates on high-frequency local features and pixel-to-pixel correlations. Thus, when these features and correlations are present in OOD data, the corresponding likelihoods are inadvertently encouraged to become large through the model's implicit bias; we hypothesize other DGMs behave similarly (see \autoref{appx:extra-likelihood-pathologies} for an extended discussion). Our work is thus complementary to that of \citet{kirichenko2020normalizing} and \citet{schirrmeister2020understanding}: even when $p_\theta$ is a good model for the true data-generating distribution, we show that $p_\theta \approx p_0$ can be violated around a set $\mathcal{M}_\theta \setminus \mathcal{M}_{\text{in}}$ of small ``volume'', whereas they provide an explanation of why this set sometimes contains $\mathcal{M}_{\text{out}}$.
 
The connection between LID and ``volume'' also explains the directionality of the paradox; i.e., why it only arises when OOD data is simpler (in that it has lower intrinsic dimension) than in-distribution data. In the non-pathological case, when $\mathcal{M}_{\text{out}}$ is more complex (i.e., higher-dimensional) than $\mathcal{M}_{\text{in}}$, assigning large densities to $\mathcal{M}_{\text{out}}$ would necessarily correspond to a higher ``volume'' and hence high probability mass. 
High probability mass would imply that $p_\theta$ generates samples $\Mout$, which of course never occurs in practice. However, when $\mathcal{M}_{\text{out}}$ is lower-dimensional than $\mathcal{M}_{\text{in}}$, the model $p_\theta$ is able to assign lower $\LID_{\theta}(\vect{x})$, and thus lower ``volume'', to $\mathcal{M}_{\text{out}}$. This allows it to simultaneously assign  pathologically large densities and vanishingly small probability mass to $\mathcal{M}_{\text{out}}$. Only in this second case can $p_\theta$ closely approximate $p_0$ while also assigning high densities to $\manifold_\text{out}$.

It follows that if $\LID_{\theta}(\vect{x})$ has a small value relative to in-distribution data, we can expect the probability mass that $p_\theta$ assigns around $\vect{x}$ to be negligible -- even if $p_\theta(\vect{x})$ is large -- suggesting that $\vect{x}$ should be classified as OOD.

\subsection{Estimating LID}

We have now justified the use of $\LID_{\theta}(\vect{x})$ for OOD detection, yet this quantity cannot be evaluated, only estimated, which we now show how to do.

\paragraph{LID for NFs} Consider a smooth map $f: \mathcal{Z} \rightarrow \mathcal{X}$ between two manifolds and a point $\vect{z} \in \mathcal{Z}$. If $f$ has constant rank in an open neighbourhood around $\vect{z}$, then the intrinsic dimension of $\vect{x} = f(\vect{z})$ on its image is given by the rank of the differential of $f$ at $\vect{z}$. When $f$ is a NF, $\LID_\theta(\vect{x})$ is thus formally given by $\rank \rmJ(\vect{z}) $ \citep{horvat2022}. Technically, NFs have full rank Jacobians by construction, making $\LID_\theta(\vecx) = d$ for all $\vecx$. However, since NFs concentrate mass around the low-dimensional $\mathcal{M}_\theta$ they are not numerically invertible \citep{cornish2020relaxing, behrmann2021understanding}, and so numerically they assign LIDs of less than $d$ to most points. For a given NF and a query $\vect{x}$, we thus estimate the corresponding (numerical) LID as
\begin{equation}\label{eq:lid_nf}
    \widehat{\LID}_\theta^{\text{NF}}(\vect{x}) \coloneqq \left\vert\{i \in [d]: \sigma_i^{\text{NF}}(\vect{x}) > \tau\}\right\vert,
\end{equation}
where $[d] = \{1, \ldots, d\}$, $\sigma_i^{\text{NF}}(\vect{x})$ is the $i$-th singular value of $\rmJ(\vect{z})$, and $\tau > 0$ is a threshold hyperparameter specifying which singular values are numerically equal to zero. 

\paragraph{LID for DMs} As previously mentioned, \citet{stanczuk2022your} developed an LID estimator for variance exploding DMs (\autoref{eq:lid_stan}) which is based on $s_\theta(\vect{x}', t_0)$ orthogonally pointing towards $\mathcal{M}_\theta$. We found better performance with \emph{variance preserving} DMs (\aref{appx:vesde_vs_vpsde}), where $h(\vect{x}, t) = -\tfrac{1}{2}\beta(t)\vect{x}$ and $g(t) = \sqrt{\beta(t)}$ for an affine function $\beta: [0, T] \rightarrow \R_{>0}$. In this case the direction of the drift in \autoref{eq:backward_sde_exact} is not given by $s_{T-t}(\vect{y}_t)$ anymore, but by $s_{T-t}(\vect{y}_t) + \tfrac{1}{2}\vect{y}_t$ instead. Accordingly, we modify $\rmS(\vect{x}) \in \R^{d \times k}$, where $k > d$, as
\begin{equation}
    \rmS(\vect{x}) = \Bigg[ s_\theta(\vect{x}^1_{t_0}, t_0)+\dfrac{\vect{x}^1_{t_0}}{2} \Bigg\vert \cdots \Bigg\vert s_\theta(\vect{x}^k_{t_0}, t_0)+\dfrac{\vect{x}^k_{t_0}}{2}\Bigg],
\end{equation}
whose columns we now expect to point orthogonally towards $\mathcal{M}_\theta$.\footnote{While intuitive, adding $\vect{x}_{t_0}^j$ to the $j$-th column of $\rmS(\vect{x})$ is an ad-hoc modification to the estimator proposed by \citet{stanczuk2022your} to account for our use of variance preserving DMs. In practice this modification does not drastically affect the corresponding LID estimate, as numerically it is similar to adding the same constant vector to every column.} Similarly to NFs, $\rank \rmS(\vect{x})$ can technically match $d$ even though many of its singular values are almost zero. We thus estimate the (numerical) LID of DMs as
\begin{equation}\label{eq:lid_dm}
    \widehat{\LID}_\theta^{\text{DM}}(\vect{x}) \coloneqq d - \left\vert\{i \in [d] : \sigma_i^{\text{DM}}(\vect{x}) > \tau\}\right\vert,
\end{equation}
where $\sigma_i^{\text{DM}}(\vect{x})$ is the $i$-th singular value of $\rmS(\vect{x})$.

\paragraph{Setting the threshold} Both LID estimators presented above require setting $\tau$ to threshold singular values. We found that in practice, no single value of $\tau$ performed well across all datasets. To avoid having to manually tune this hyperparameter, we propose a way to set $\tau$ only using the available (in-distribution) training data. Specifically, we leverage local principal component analysis (LPCA) \citep{fukunaga1971algorithm}, which is a fast and simple LID estimator. 
Roughly, for a given in-distribution $\vect{x}$ and a provided dataset, LPCA uses the nearest neighbours of $\vect{x}$ in the dataset to construct a matrix whose rank approximates the LID at $\vect{x}$, and we calibrate $\tau$ to match the LPCA estimates. We reiterate that estimators based on nearest neighbours such as LPCA are not directly useful for identifying OOD data, since they require local samples around a query which are unavailable for OOD data and cannot be generated by $p_\theta$. For a detailed evaluation of the proposed LID estimators, see \autoref{appx:lid-estimator-eval}.

\subsection{Putting it All Together}\label{sec:all_together}

So far we have argued that LID can be used for OOD detection and have shown how to obtain estimates $\widehat{\LID}_\theta(\vect{x})$ by using \autoref{eq:lid_nf} or \autoref{eq:lid_dm}. In summary, three mutually exclusive cases can happen for a point $\vect{x}$: $(i)$ $\log p_\theta(\vect{x})$ is small (relative to in-distribution data). $(ii)$ $\log p_\theta(\vect{x})$ is large and $\widehat{\LID}_\theta(\vect{x})$ is small. In both of these cases $p_\theta$ assigns negligible probability mass around $\vect{x}$, which in turn means $\vect{x}$ should be classified as OOD. $(iii)$ $\log p_\theta(\vect{x})$ and $\widehat{\LID}_\theta(\vect{x})$ are both large, in which case the likelihood spiking on $\vect{x}$ is not pathological, implying that $\vect{x}$ should be classified as in-distribution. This leads to our proposed dual threshold OOD detection method, described in \autoref{alg:dual_threold}, where $\psi_\mathcal{L}$ and $\psi_{\LID}$ are the log-likelihood and LID thresholds, respectively. We highlight that our method differs from standard (single threshold) likelihood-based OOD detection only in that we classify case $(ii)$ as OOD instead of in-distribution.

\begin{algorithm}[t]
\caption{Dual threshold OOD detection, returns $\texttt{True}$ if $\vect{x}$ is deemed OOD, and $\texttt{False}$ if deemed in-distribution.}
\label{alg:dual_threold}
\begin{algorithmic}[1]
\Require $\log p_\theta(\vect{x}), \widehat{\LID}_\theta(\vect{x}), \psi_\mathcal{L}, \psi_\LID$
    \If{$\log p_\theta(\vect{x}) < \psi_\mathcal{L}$}
        \State \Return $\texttt{True}$ \Comment{case $(i)$}
    \EndIf
    \If{$ \widehat{\LID}_\theta(\vect{x}) < \psi_\LID$}
        \State \Return $\texttt{True}$ \Comment{case $(ii)$}
    \EndIf
    \State \Return $\texttt{False}$ \Comment{case $(iii)$}
\end{algorithmic}
\end{algorithm}

\section{Related Work} \label{sec:related_work}

A substantial amount of research into likelihood pathologies tries to explain the underlying causes of the OOD paradox. One particular line of research proposes probabilistic explanations: \citet{choi2018waic} and \citet{nalisnick2019detecting} put forth the ``typical set'' hypothesis, which has been contested in follow-up work. For example, \citet{le2021perfect} argue that likelihood rankings not being invariant to data reparameterizations causes the paradox, whereas \citet{caterini2022entropic} claim it is the lower entropy of ``simpler'' distributions as compared to the higher entropy of more ``complex'' ones -- which somewhat aligns with our work, although we use intrinsic dimension instead of entropy to quantify complexity. 

We diverge from these explanations in that they all assume, sometimes implicitly, that $\mathcal{M}_{\text{in}}$ and $\mathcal{M}_{\text{out}}$ overlap. We find it extremely plausible, for example, that the intersection between CIFAR10 and SVHN images is empty. In this sense, we are more in agreement with \citet{zhang2021understanding}, who propose a similar explanation to ours based on probability mass. Nonetheless, we differ from the work of \citet{zhang2021understanding} in several key ways: $(i)$ they blame poor model fit as the culprit, which is inconsistent with our results showing that likelihoods do not distinguish between OOD and generated samples; $(ii)$ they do not establish a connection to LID; and $(iii)$ they do not empirically verify their explanation since they do not propose a method to address the issue.

Another line of work aims to build DGMs which do not experience the OOD paradox \citep{li2022out}, sometimes at the cost of generation quality. For example, \citet{grathwohl2019your} and \citet{liu2020energy} argue that the training procedure of energy-based models (EBMs) \citep{xie2016theory, du2019implicit} provides inductive biases which help OOD detection, and \citet{yoon2021autoencoding} construct an EBM specifically designed for this task. \citet{kirichenko2020normalizing} and \citet{loaiza2022diagnosing} first embed data into semantically rich latent spaces, and then employ dense neural network architectures, thus minimizing susceptibility to local high-frequency features.  We differ from these works in that they attempt to build DGMs that are better at likelihood-based OOD detection, whereas we only leverage a pre-trained model.

Other works use ``outside help'' or auxiliary models. Some methods assume access to an OOD dataset \citep{nalisnick2019detecting}, require class labels \citep{gornitz2013toward,  ruff2020deep, van2021feature, liu2022simple}, or leverage an image compression algorithm \citep{serra2019input}. Some other works, while fully unsupervised, require training auxiliary models on distorted data \citep{ren2019likelihood}, on the incoming test datapoint with regularization \citep{xiao2020likelihoodregret}, or on data summary statistics \citep{morningstar2021density}. A final line of work leverages DMs for OOD detection. \citet{graham2023denoising}, \citet{10.5555/3618408.3619344}, and \citet{choi2023projection} propose methods based on reconstruction errors. \citet{goodier2023likelihood}, who use the variational formulation of DMs \citep{sohl2015deep, ho2020denoising} rather than the score-based one, adopt a likelihood ratio approach which averages the DM loss across various noise levels. Again, we are different from these works in that we do not just care about empirical performance, but also about understanding why likelihoods alone fail -- and leveraging this understanding for fully unsupervised OOD detection.
 
\section{Experiments} \label{sec:experiments}
\begin{figure*}[t]\captionsetup{font=footnotesize}
    \centering

    \begin{minipage}[t]{\textwidth}
    \begin{subfigure}[t]{0.48\textwidth}
    \centering
        \includegraphics[scale=0.225,angle=90]{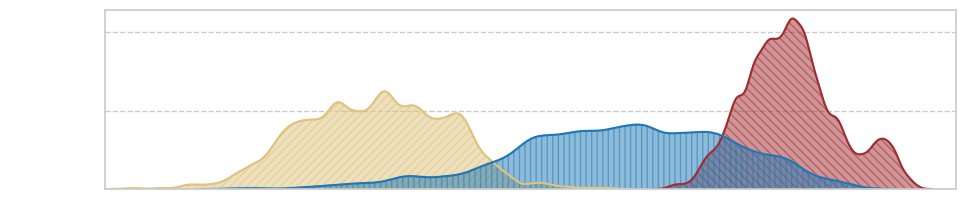}
        \includegraphics[scale=0.23]{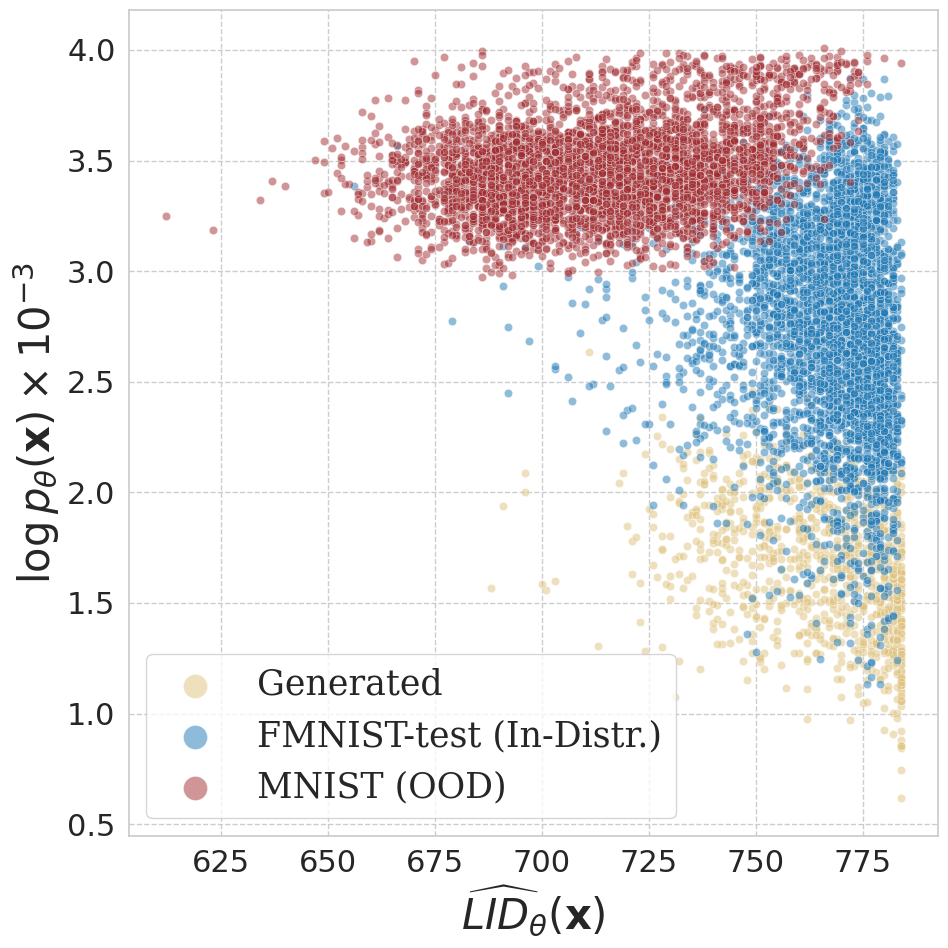}\\
        \hspace{38pt}
        \includegraphics[scale=0.22]{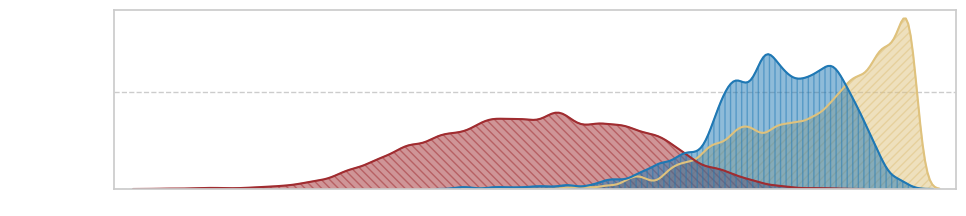}
        \caption{}
    \end{subfigure}
    \begin{subfigure}[t]{0.48\textwidth}
        \centering
        \includegraphics[scale=0.23]{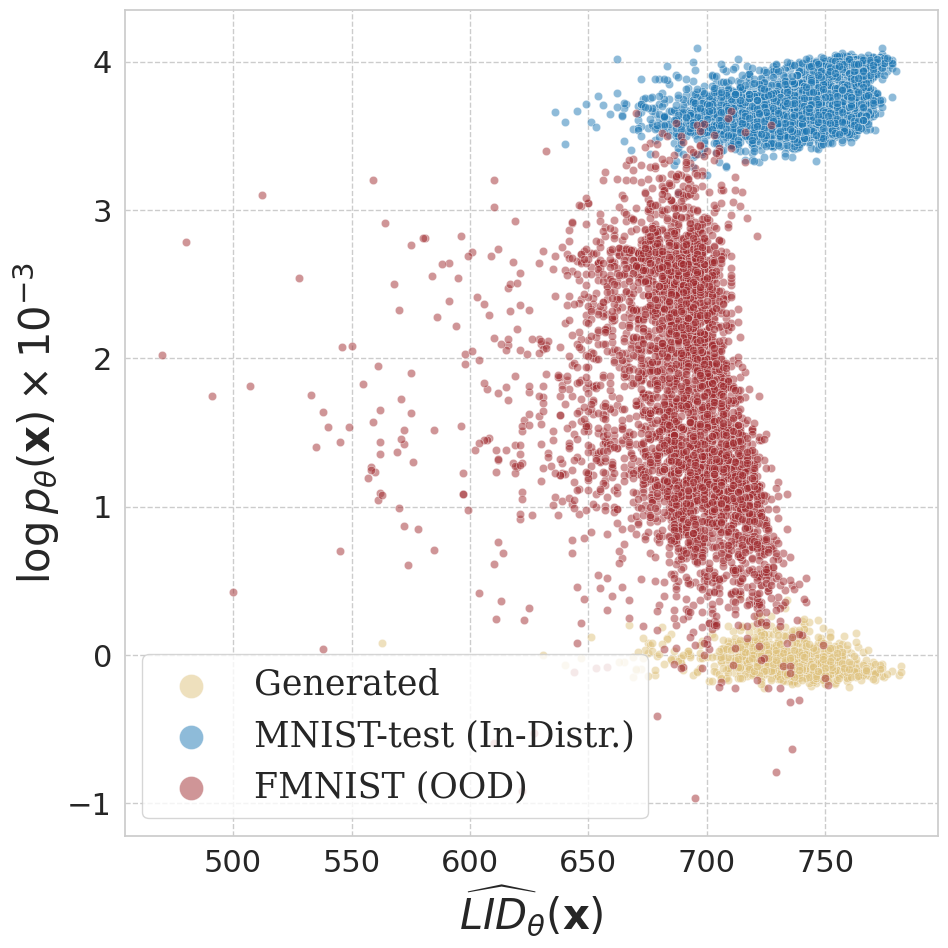}
        \scalebox{-1}[1]{\includegraphics[scale=0.225,angle=90]{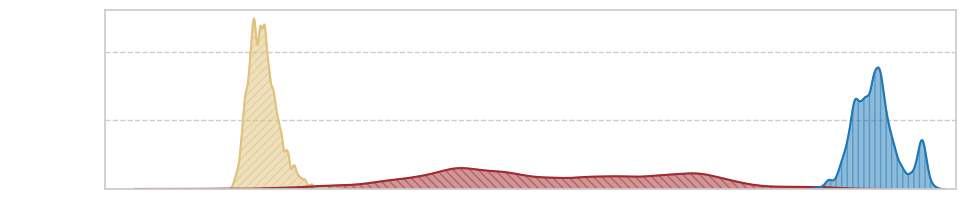}}\\
        \hspace{-34pt}
        \includegraphics[scale=0.22]{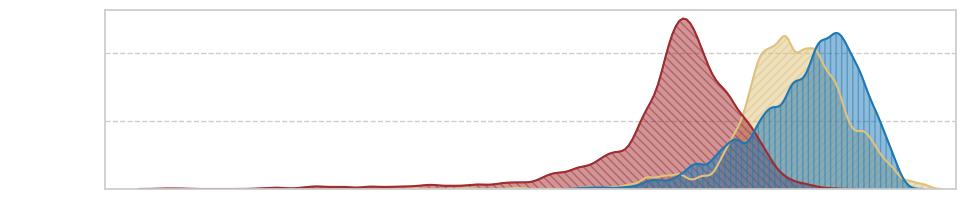}
        \caption{}
        \end{subfigure}
    \end{minipage}
    
    \captionsetup{belowskip=-10pt}
    \caption{
    LID estimates and likelihood scatterplots, along with corresponding marginals. \textbf{(a)} FMNIST-trained model, evaluated on FMNIST, MNIST, and generated samples. \textbf{(b)} MNIST-trained model, evaluated on FMNIST, MNIST, and generated samples.}
    \label{fig:fmnist_mnist_scatterplots}
\end{figure*}

\paragraph{Setup} We compare datasets within two classes: $(i)$ $28\times 28$ greyscale images, including FMNIST, MNIST, Omniglot \citep{lake2015human}, and EMNIST \citep{cohen2017emnist}; and $(ii)$ RGB images resized to $32\times 32 \times 3$, comprising SVHN, CIFAR10 and CIFAR100 \citep{krizhevsky2009learning}, Tiny ImageNet \citep{le2015tiny}, and a simplified, cropped version of CelebA \citep{kist_andreas_m_2021_5561092}. 
We give experimental details on model training in \aref{appx:flow_hp_setting} and \aref{appx:diffusion_hp_setting}. Due to space constraints, we report results for all dataset pairs in \aref{appx:all_aucs_gen_pathologies}. 

\paragraph{Evaluation} 
OOD detection methods use the area under the curve (AUC) of the receiver operator characteristic (ROC) curve for evaluation. The true positive rate (TPR) and false positive rate (FPR) from an OOD classifier correspond to points on the FPR-TPR plane. 
By sweeping over all possible threshold values, these points determine the ROC graph. 
For single threshold OOD classifiers, the graph provides points on a curve indicating the best achievable TPR for each FPR, the area under which is denoted as AUC-ROC.
On the other hand, the ROC graph for dual threshold classifiers corresponds to points on a surface -- not a curve -- on the FPR-TPR plane. The upper boundary of this surface defines a curve, which also indicates the best achievable TPR for each FPR. Thus, in a slight abuse of terminology, we also denote the area under this curve as AUC-ROC, as it is directly comparable to that of single threshold classifiers. See \aref{appx:optimal_roc_curves} for a thorough explanation of such ROC curves and how we compute their AUC.

\paragraph{LID with Likelihoods Isolates OOD Regions} We compute log-likelihoods and LID estimates for NFs trained on FMNIST and MNIST, with results shown in \autoref{fig:fmnist_mnist_scatterplots}.  
The scatterplots with both likelihoods and LIDs show clear separation between in-distribution and OOD, despite the likelihood and LID marginals overlapping. Furthermore, the ``directions'' predicted by our method are correct: in the pathological case (FMNIST-trained), we see that while OOD points have higher likelihoods, they also have lower LIDs; whereas in the non-pathological case (MNIST-trained), likelihoods are lower for OOD data. These results highlight not only the importance of using LID estimates for OOD detection, but also that of combining them with likelihoods, as the two together provide a proxy for probability mass.

\begin{figure*}\captionsetup{font=footnotesize}
    \centering
    \begin{minipage}[b]{\textwidth}
        \centering
        \begin{subfigure}[b]{0.245\textwidth}
            \includegraphics[width=\textwidth]{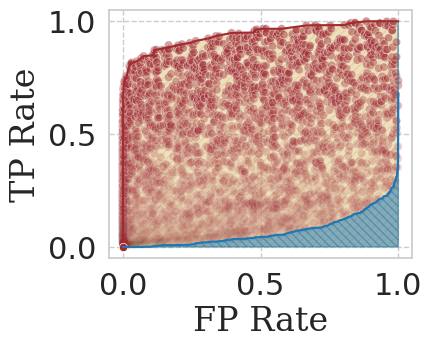}
            \caption{~\textbf{FMNIST~vs.~MNIST}:\\AUC-ROC boost ($0.070 \to 0.953$)}
            \label{fig:fmnist_vs_mnist_ROG}
        \end{subfigure}
        \hfill 
        \begin{subfigure}[b]{0.245\textwidth}
            \includegraphics[width=\textwidth]{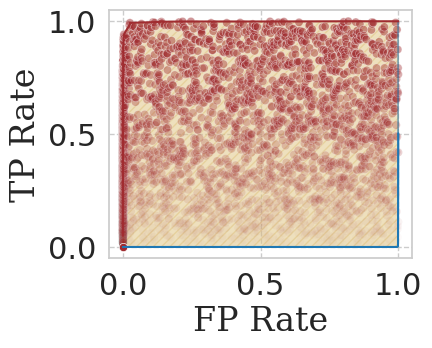}
            \caption{~\textbf{FMNIST-gen~vs.~Omniglot}:\\AUC-ROC boost ($0.000 \to 0.996$)}
            \label{fig:fmnist_vs_omniglot_ROG}
        \end{subfigure}
        \hfill 
        \begin{subfigure}[b]{0.245\textwidth}
            \includegraphics[width=\textwidth]{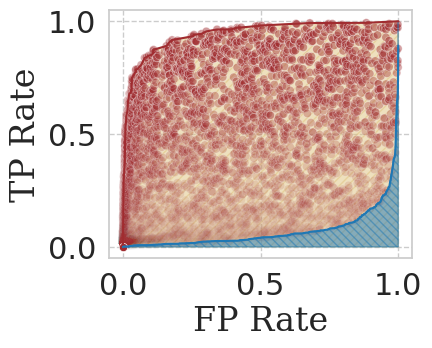}
            \caption{~\textbf{CIFAR10~vs.~SVHN}:\\ AUC-ROC boost ($0.060 \to 0.926$)}
            \label{fig:cifar10_vs_svhn_ROG}
        \end{subfigure}
        \hfill 
        \begin{subfigure}[b]{0.245\textwidth}
            \includegraphics[width=\textwidth]{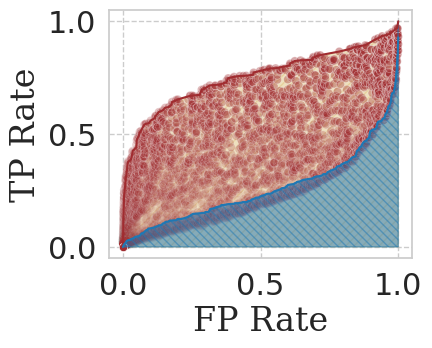}
            \caption{\textbf{CIFAR10-gen~vs.~CelebA}:\\AUC-ROC boost ($0.258 \to 0.733$)}
            \label{fig:cifar10_vs_celeba_ROG}
        \end{subfigure}
    \end{minipage}
    \captionsetup{belowskip=-10pt}
    \caption{
    ROC visualizations for select pathological OOD tasks on NFs. The red dots correspond to the FPR-TPR pairs of our method obtained from different dual thresholds, the yellow areas correspond to the region under the associated Pareto frontier (i.e.\ the upper boundary of the red dots), while the blue areas represent the region below the ROC curve for single threshold likelihood-based classifiers. \textbf{(a)} FMNIST-trained model with MNIST as OOD; \textbf{(b)} as in (a) except we now discern between generated samples and MNIST. \textbf{(c)} CIFAR10-trained model with SVHN as OOD; \textbf{(d)} as in (c) except we now discern between generated samples and CelebA.}
    \label{fig:ROG_ROC_AUC}
    \vspace{5pt}
\end{figure*}

\paragraph{Visualizing the Benefits of Dual Thresholding} The separation of OOD and in-distribution data shown in the scatterplots in \autoref{fig:fmnist_mnist_scatterplots} confirms that likelihood/LID pairs contain the needed information for OOD detection. However, it remains to show that \autoref{alg:dual_threold} succeeds at this task (recall that we cannot simply train a classifier to differentiate between red and blue points in \autoref{fig:fmnist_mnist_scatterplots} since the red OOD points are unavailable when designing the OOD detector). \autoref{fig:ROG_ROC_AUC} provides a visual comparison showcasing the ROC curves from our dual thresholding technique versus the ROC curves constructed by single threshold classifiers using only likelihoods. 
These results show a dramatic boost in AUC-ROC performance across four different pathological scenarios, highlighting the relevance of combining likelihoods with LIDs for OOD detection. For further experiments, please refer to the ablations in \aref{appx:ablations}. 

\begin{table*}[t]\captionsetup{font=footnotesize}
\vspace{15pt}
\centering
\footnotesize
\caption{AUC-ROC (higher is better). The top part of the table contains NF-based approaches, the middle rows contain DM-based approaches, and the last row shows an EBM-based one. \textbf{Notation}: $^\ast$ indicates tasks where likelihoods alone do not exhibit pathological behaviour; $\ddagger$ indicates methods that employ external information or auxiliary models. For the NF and DM methods, we bold the best performing approach among themselves, and the EBM model is bolded when it surpasses all others.}
\begin{tabularx}{\textwidth}{l*{8}{Y}}
\toprule
~~~Trained on & \multicolumn{2}{c}{MNIST}$^\ast$ & \multicolumn{2}{c}{FMNIST} & \multicolumn{2}{c}{CIFAR10} & \multicolumn{2}{c}{SVHN}$^\ast$ \\
\cmidrule(r){2-3} \cmidrule(lr){4-5} \cmidrule(lr){6-7} \cmidrule(l){8-9}
OOD Dataset& {\scriptsize FMNIST} & {\scriptsize Omniglot} & {\scriptsize MNIST} & {\scriptsize Omniglot} & {\scriptsize SVHN} & {\scriptsize CelebA} & {\scriptsize CIFAR10} & {\scriptsize CelebA} \\
\midrule
NF Likelihood & $\mathbf{1.000}$ & $0.796$ & $0.073$ & $0.085$ & $0.063$ & $0.391$ & $\mathbf{0.987}$ & $\mathbf{0.996}$ \\
NF $\Vert\frac{\partial}{\partial \vect{x}} \log p_\theta(\vect{x})\Vert_2$& $0.156$ & $ 0.444$ & $0.516$ & $0.538$ & $0.722$ & $0.433$ & $0.200$ & $0.080$ \\
Complexity Correction$^\ddagger$ & $0.945$ & $0.852$ & $0.939$ & $\mathbf{0.935}$ & $0.835$ & $0.479$ & $0.771$ & $0.639$ \\ 
NF Likelihood Ratios$^\ddagger$ & $0.944$ & $0.722$ & $0.666$ & $0.639$ & $0.299$ & $0.396$ & $0.302$ & $0.099$ \\
NF Dual Threshold (Ours)& $\mathbf{1.000}$ &$\mathbf{0.855}$ & $\mathbf{0.951}$ & $0.864$ & $\mathbf{0.936}$ & $\mathbf{0.655}$ & $\mathbf{0.987}$ & $\mathbf{0.996}$ \\
\midrule
DM Likelihood & $0.996$ & $\mathbf{1.000}$ & $0.240$ & $0.952$ & $0.064$ & $0.360$ & $\mathbf{0.996}$ & $\mathbf{0.996}$ \\
DM $\Vert s_\theta(\vect{x}, 0)\Vert_2$& $0.919$ & $0.004$ & $0.075$ & $0.001$ & $0.883$ & $\mathbf{0.716}$ & $0.120$ & $0.145$ \\
DM Reconstruction$^\ddagger$ & $\mathbf{1.000}$ & $0.999$ & $\mathbf{0.970}$ & $\mathbf{0.992}$ & $0.876$ & $0.630$ & $0.984$ & $0.995$\\ 
DM Likelihood Ratios & $0.224$ & $0.296$ & $0.781$ & $0.388$ & $0.829$ & $0.553$ & $0.326$ & $0.357$\\
DM Dual Threshold (Ours)& $0.996$ & $\mathbf{1.000}$ & $0.912$ & $0.959$ & $\mathbf{0.944}$ & $0.648$ & $\mathbf{0.996}$ & $\mathbf{0.996}$ \\
\midrule
NAE& $\mathbf{1.000}$ & $0.994$ & $\mathbf{0.995}$ & $0.976$ & $0.919$ & $\mathbf{0.887}$ & $0.948$ & $0.965$ \\
\bottomrule
\end{tabularx}
\label{tab:results-table}
\end{table*}
\paragraph{Quantitative Comparisons} In the top part of \autoref{tab:results-table}, we compare our method against several normalizing flow (NF) baselines, all of which are evaluated using the exact same pre-trained NF as our method. These baselines are: $(i)$ na\"ively labelling large log-likelihoods $\log p_\theta(\vect{x})$ as in-distribution, which strongly fails at identifying ``simpler'' distributions as OOD when trained on ``complex'' datasets; $(ii)$ using $\Vert \frac{\partial}{\partial \vect{x}}\log p_\theta(\vect{x}) \Vert_2$ as a proxy for local probability mass as proposed by \citet{grathwohl2019your}, which performs inconsistently across tasks; $(iii)$ the complexity correction method of \citet{serra2019input}, which uses image compression information to adjust the inflated likelihood observed in OOD datapoints -- despite this comparison being unfair in that the baseline accessed image compression algorithms in addition to the NF, we beat it across all tasks except one; $(iv)$ 
the likelihood ratios approach of \citet{ren2019likelihood}, which is once again unfair as it employs an auxiliary likelihood-based reference model to compute ratios, yet we uniformly beat it across tasks.

Besides the strong empirical performance of our method with NFs, other aspects of the top part of \autoref{tab:results-table} warrant attention. Both the complexity correction and likelihood ratio baselines lose performance over na\"ively using likelihoods on non-pathological tasks, i.e.\ when models are trained on relatively ``simple'' data like MNIST or SVHN. Since likelihoods perform well at these tasks, they are often considered ``easy'' and thus omitted from comparisons. The fact that these baselines struggle at these tasks is a novel finding that suggests these methods ``overfit'' to the pathological tasks. See \aref{appx:baselines} for a thorough discussion. 

Additionally, we test our dual threshold method with diffusion models (DMs) in the middle rows of \autoref{tab:results-table}, and compare its performance against: $(i)$ using only likelihoods, which fails at most pathological tasks; $(ii)$ using the norm of the likelihood derivative, or equivalently, the norm of the score function \citep{grathwohl2019your} -- this also fails in pathological tasks; $(iii)$ the reconstruction-error-based approach of \citet{graham2023denoising}, which provides the strongest baseline and is very similar to \citet{choi2023projection} -- we find it noteworthy that we perform on par with this baseline, beating it at five out of eight tasks, despite the fact that it is not fully unsupervised as it relies on a pre-trained LPIPS encoder \citep{zhang2018unreasonable}; $(iv)$ the likelihood ratios method of \citet{goodier2023likelihood}, which does not perform well.\footnote{Note that \citet{graham2023denoising} applied their method on latent diffusion models \citep{rombach2022high}, which, as discussed in \autoref{sec:related_work}, makes OOD detection easier. We also point out that \citet{goodier2023likelihood} used the variational formulation of DMs \citep{ho2020denoising}, and we adapted their method to score-based models. These discrepancies explain the differences between the numbers in \autoref{tab:results-table} and those reported in these papers.} We used the exact same DM for every comparison, and find that our method outperforms all fully unsupervised baselines that do not leverage outside data.

Overall, we believe it is remarkable that our dual threshold outperforms every baseline for NFs and performs on par with the strongest DM baseline, both pathological and non-pathological tasks, despite some baselines having access to additional information. We see these results as strong evidence supporting the understanding that we derived about the OOD paradox and its connection to LID. We also highlight that, as mentioned in \autoref{sec:background}, we identified cases of likelihoods behaving pathologically on generated samples. In \aref{appx:all_aucs_gen_pathologies}, we show that our dual threshold method also excels at detecting these scenarios.

The last row of \autoref{tab:results-table} shows normalized autoencoders (NAEs) \citep{yoon2021autoencoding}. NAEs are EBMs specially tailored for OOD detection at the cost of generation quality, but to the best of our knowledge achieve state-of-the-art performance on fully unsupervised, likelihood-based OOD detection. Once again, we believe that the empirical results of our dual threshold method are remarkable: we achieve similar performance to NAEs on most tasks, even outperforming them on four, despite using a general purpose model $p_\theta$, not one explicitly designed for OOD detection.

\vspace{-2mm}
\section{Conclusions, Limitations, and Future Work} \label{sec:conclusion}
In this paper we studied the OOD detection paradox, where likelihood-based DGMs assign high likelihoods to OOD points from ``simpler'' datasets, but do not generate them. We proposed a geometric explanation of how the paradox can arise as a consequence of models assigning low probability mass around these OOD points when they have small intrinsic dimensions. We then leveraged LID estimators for our dual threshold OOD detection method. Having decidedly outperformed the use of likelihoods by themselves, our results strongly support our geometric explanation. We believe that extending the utility of this geometric viewpoint and of LID beyond OOD detection is an extremely interesting path for follow-up work.

We highlight that the LID estimator of \citet{stanczuk2022your}, which we heavily relied upon for DMs, obtains very different estimates on image datasets than previously established LID estimators. Our dual threshold technique works despite this discrepancy likely because only LID rankings are relevant for OOD detection. Further improving DM-based LID estimators is a promising avenue to boost performance.

Finally, while our ideas are widely applicable to any density model, the current incarnation of our method is limited in that it only applies to NFs and DMs, as estimating LID is more tractable for these models. Extending our method to DGMs whose LID might be estimated as the rank of an appropriate matrix (like the Jacobian of a decoder), such as variational autoencoders \citep{kingma2014auto, rezende2014stochastic, Vahdat2020NVAEAD}, injective NFs \citep{brehmer2003flows, caterini2021rectangular, ross2021tractable}, or DMs on latent space \citep{vahdat2021score, rombach2022high} is also likely to work. Nonetheless, we see extending our method to EBMs, which achieve state-of-the-art likelihood-based OOD detection, to be a particularly promising direction for future research.

\section*{Impact Statement}

The goal of this work is to advance the understanding and methodology in the detection of out-of-distribution data using a pre-trained deep generative model. Our contribution is primarily of a theoretical nature and therefore has limited broader societal impacts on its own. However, the generality of our work lends itself to use with several existing classes of models. Therefore, we emphasize that further evaluation on diverse datasets and data modalities is required before our method is deployed as the sole tool to detect pathologies in data.

\section*{Acknowledgements}
We thank Valentin Villecroze and George Stein for useful feedback on a previous version of this manuscript. RGK acknowledges support from a Tier II Canada Research Chair and a Canada CIFAR AI Chair.

\bibliography{refs}
\bibliographystyle{icml2024}

\newpage
\appendix
\onecolumn
\section{Diagnosing Pathologies in Normalizing Flows and Diffusion Models}\label{appx:extra-likelihood-pathologies}

\begin{figure}\captionsetup{font=footnotesize}
    \centering
    \begin{minipage}[b]{\textwidth}
        \centering
        \begin{subfigure}[b]{0.48\textwidth}
            \includegraphics[width=\textwidth]{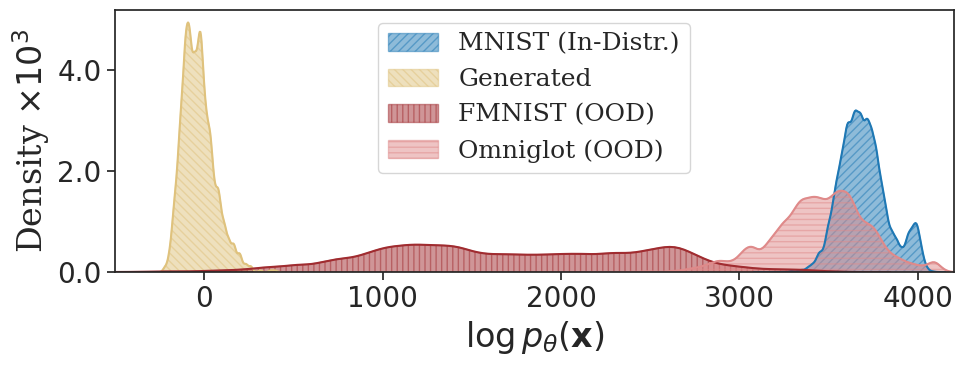}
            \caption{MNIST-trained NF vs.\ FMNIST and Omniglot}
            \label{fig:pathology_mnist_vs_fmnist_omniglot_flow}
        \end{subfigure}
        \hfill 
        \begin{subfigure}[b]{0.48\textwidth}
            \includegraphics[width=\textwidth]{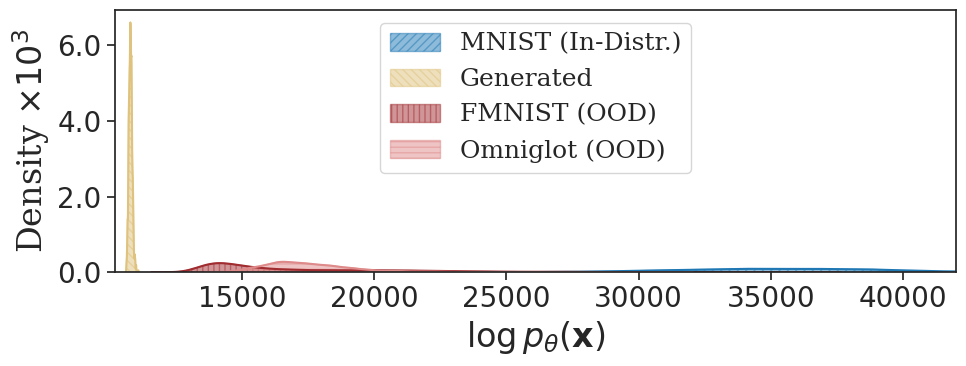}
            \caption{ MNIST-trained DM vs.\ FMNIST and Omniglot}
            \label{fig:pathology_mnist_vs_fmnist_omniglot_diffusion}
        \end{subfigure}
    \end{minipage}
    
    \vspace{1mm}
    \begin{minipage}[b]{\textwidth}
        \centering
        \begin{subfigure}[b]{0.48\textwidth}
            \includegraphics[width=\textwidth]{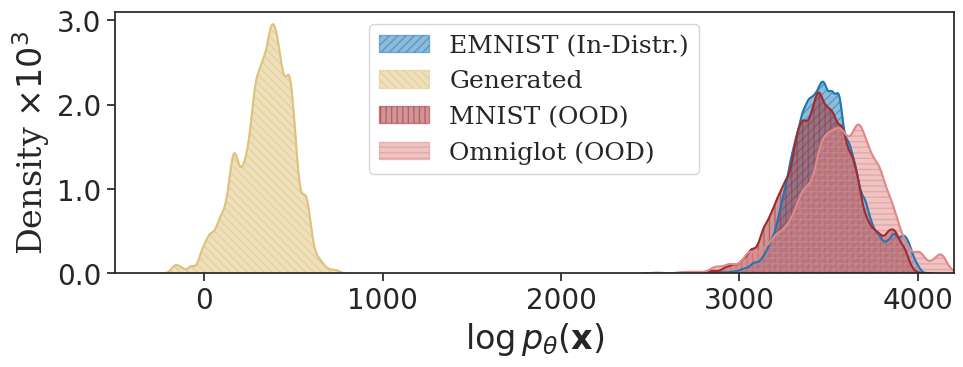}
            \caption{EMNIST-trained NF vs.\ MNIST and Omniglot}
            \label{fig:pathology_emnist_vs_mnist_omniglot_flow}
        \end{subfigure}
        \hfill 
        \begin{subfigure}[b]{0.48\textwidth}
            \includegraphics[width=\textwidth]{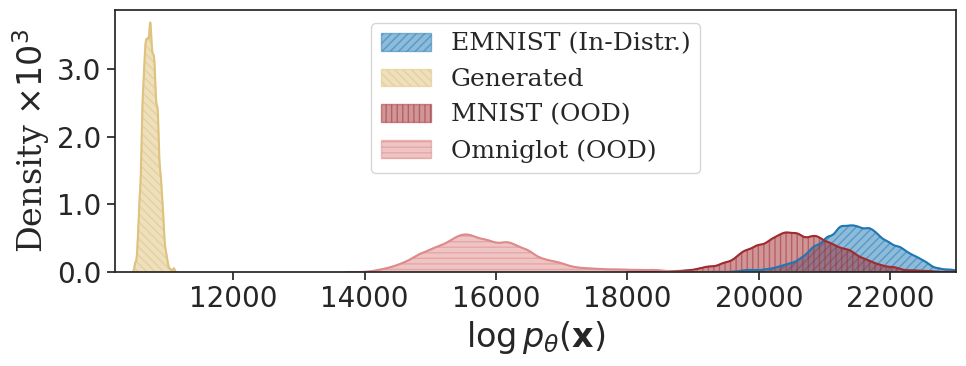}
            \caption{ EMNIST-trained DM vs.\ MNIST and Omniglot}
            \label{fig:pathology_emnist_vs_mnist_omniglot_diffusion}
        \end{subfigure}
    \end{minipage}
    
    \vspace{1mm}
    
    \begin{minipage}[b]{\textwidth}
        \centering
        \begin{subfigure}[b]{0.31\textwidth}
            \includegraphics[width=\textwidth]{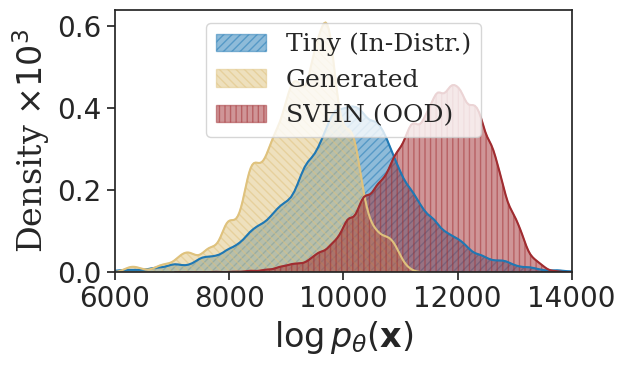}
            \caption{ Tiny-trained NF vs.\ SVHN}
            \label{fig:pathology_tiny_vs_svhn_flow}
        \end{subfigure}
        \hfill
        \begin{subfigure}[b]{0.31\textwidth}
            \includegraphics[width=\textwidth]{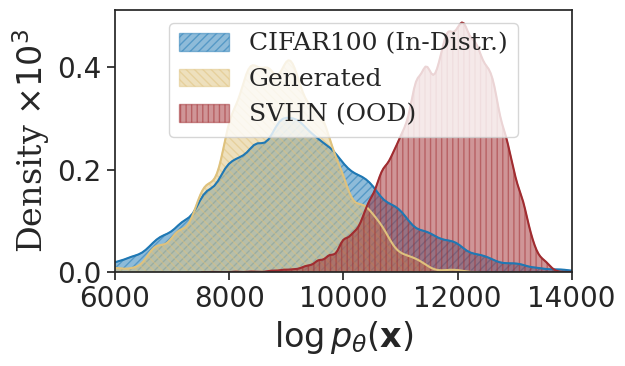}
            \caption{ CIFAR100-trained NF vs.\ SVHN}
            \label{fig:pathology_cifar100_vs_svhn_flow}
        \end{subfigure}
        \hfill
        \begin{subfigure}[b]{0.31\textwidth}
            \includegraphics[width=\textwidth]{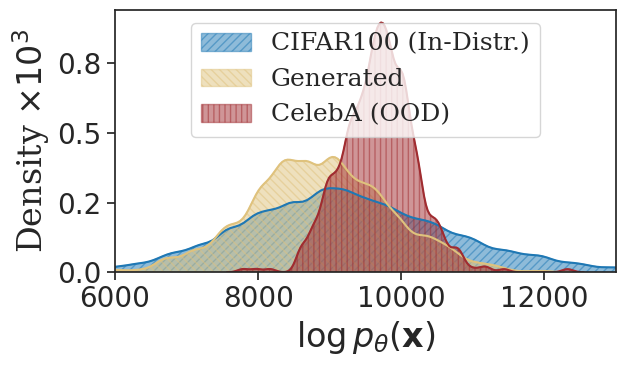}
            \caption{ CIFAR100-trained NF \textbf{vs} CelebA}
            \label{fig:pathology_cifar100_vs_celeba_flow}
        \end{subfigure}
    \end{minipage}

    \vspace{1mm}
    \begin{minipage}[b]{\textwidth}
        \centering
        \begin{subfigure}[b]{0.31\textwidth}
            \includegraphics[width=\textwidth]{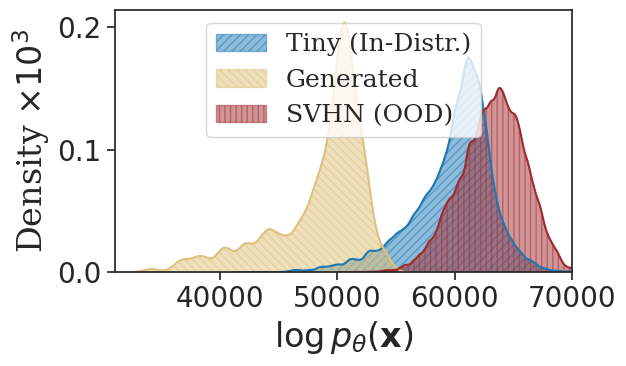}
            \caption{ Tiny-trained DM \textbf{vs} SVHN}
            \label{fig:pathology_tiny_vs_svhn_diffusion}
        \end{subfigure}
        \hfill
        \begin{subfigure}[b]{0.31\textwidth}
            \includegraphics[width=\textwidth]{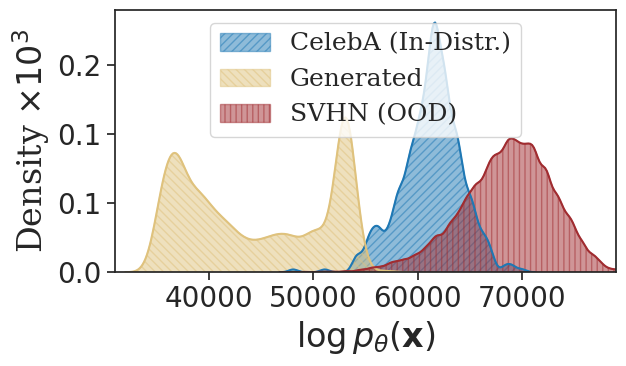}
            \caption{ CelebA-trained DM vs.\ SVHN}
            \label{fig:pathology_celeba_vs_svhn_diffusion}
        \end{subfigure}
        \hfill
        \begin{subfigure}[b]{0.31\textwidth}
            \includegraphics[width=\textwidth]{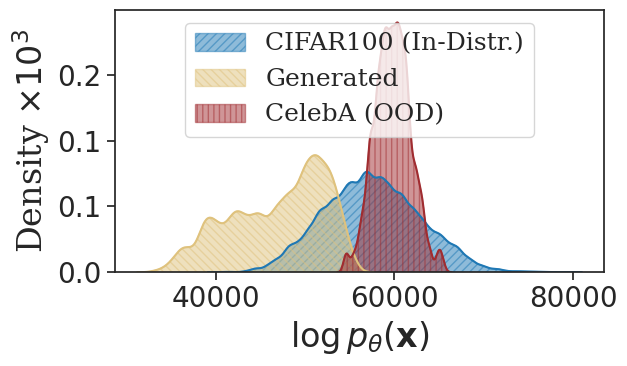}
            \caption{ CIFAR100-trained DM vs.\ CelebA}
            \label{fig:pathology_cifar100_vs_celeba_diffusion}
        \end{subfigure}
    \end{minipage}

    \vspace{1mm}
    \begin{minipage}[b]{\textwidth}
        \centering
        \begin{subfigure}[b]{0.31\textwidth}
            \includegraphics[width=\textwidth]{new_figures/cifar10_vs_svhn_diffusion.png}
            \caption{CIFAR10-trained DM vs.\ SVHN}
            \label{fig:pathology_cifar10_vs_svhn_diffusion_app}
        \end{subfigure}
        \hfill
        \begin{subfigure}[b]{0.31\textwidth}
            \includegraphics[width=\textwidth]{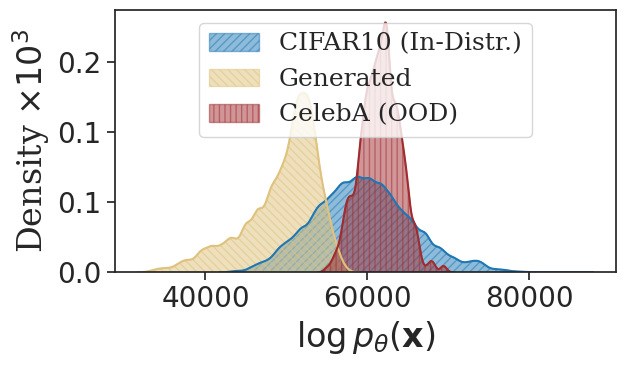}
            \caption{ CIFAR10-trained DM vs.\ CelebA}
            \label{fig:pathology_cifar10_vs_celeba_diffusion_app}
        \end{subfigure}
        \hfill
        \begin{subfigure}[b]{0.31\textwidth}
            \includegraphics[width=\textwidth]{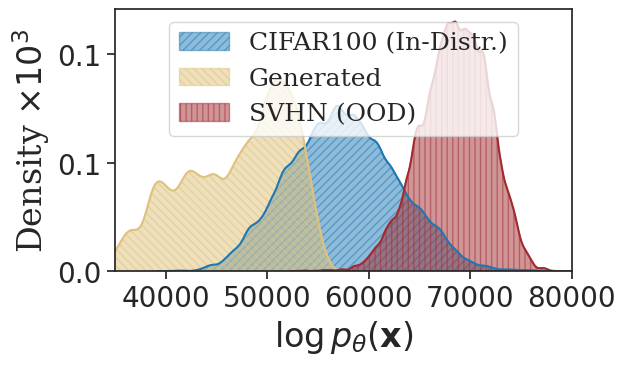}
            \caption{ CIFAR100-trained DM vs.\ SVHN}
            \label{fig:pathology_cifar100_vs_svhn_diffusion_app}
        \end{subfigure}
    \end{minipage}
    
  \captionsetup{belowskip=-11pt}
    \caption{Overview of likelihood pathologies: \textbf{(a-d)} Models trained on EMNIST or MNIST assign the highest likelihoods to in-distribution data as expected, but obtain strikingly low likelihoods on generated samples -- even lower than for OOD data. \textbf{(e-m)} Pathologies on RGB datasets where both the in-distribution samples and the generated samples are assigned likelihoods smaller than that of OOD datapoints.}
    \label{fig:extra_likelihood_pathologies}
\end{figure}

In this section, we list the full extent of the pathologies we identified in our experiments. 
The first class is the standard one, in which models assign equal or higher likelihoods to OOD data than to in-distribution data; furthermore, we observe a new class of pathologies where the model assigns low likelihoods to its own generated samples compared to OOD data. In addition to \autoref{fig:likelihood_pathologies}, which shows that FMNIST vs.\ MNIST, CIFAR-10 vs.\ SVHN, and CelebA vs.\ SVHN are pathological, \autoref{fig:extra_likelihood_pathologies} depicts pathological behaviour for EMNIST vs.\ MNIST, EMNIST vs.\ Omniglot, Tiny ImageNet vs.\ SVHN, CelebA vs.\ SVHN, CIFAR100 vs.\ SVHN, and CIFAR100 vs.\ CelebA. 

Regarding the second class of pathologies,  we observe a stark difference between the likelihoods of generated samples and the in-distribution ones. To demonstrate this, we also visualize the likelihoods of generated samples in \autoref{fig:likelihood_pathologies} and \autoref{fig:extra_likelihood_pathologies}. Notably, generated sample likelihoods are almost always smaller than both in-distribution as well as OOD samples. To the best of our knowledge, generated samples having lower likelihoods than OOD data is a new class of pathologies not previously discussed in the literature. This raises new unexplained phenomena, even in cases such as MNIST- and Omniglot-trained models which were previously thought to be non-pathological \citep{nalisnick2018deep}.

The rationale of \citet{schirrmeister2020understanding} might provide an explanation for why these new pathologies occur in NFs. They claim that the multi-scale NF architectures used for modelling images pick up on high-frequency features, such as sharp edges, which are prevalent in any natural dataset; this prompts the latent variables corresponding to shallow scales to \textit{sharply} center around zero (i.e.\ with a very small variance), and in turn the likelihood of these latent variables strongly inflates the total likelihood, regardless of whether the original datapoint is OOD or not. When passing a generated sample (that appears semantically similar to in-distribution data) inversely through an NF, we get shallow-scale latent variables that have a standard deviation of $1$ by design. Therefore, compared to OOD data selected from natural images, these latent variables are not as sharply concentrated around $0$, and hence produce relatively smaller likelihoods. That said, none of the explanations in related works such as those by \citet{kirichenko2020normalizing} or \citet{schirrmeister2020understanding} directly provide any conclusive insight into the inductive biases that make DMs behave this way. 
Among related work on DMs, \citet{goodier2023likelihood} offers a potential explanation for the pathology, suggesting that the score network prioritizes high-frequency features in earlier timesteps—a characteristic also common in OOD data; however, the evidence they present does not adequately address the occurrence of the pathology on generated samples, which is a novel observation of our work.

We emphasize that these explanations clarify why NF or DM likelihoods for OOD data points can behave pathologically, but not why low likelihood data is generated in the first place. Therefore they do not contradict, but rather complement, our explanation. 
Similar to in-distribution data, generated samples have comparatively higher probability masses, potentially even surpassing that of in-distribution data. 
Consequently, in scenarios where the likelihood is pathological for OOD data, the LID is anticipated to be small; our experiments in \autoref{tab:results-table-gen} further substantiate this observation by getting consistent performance across OOD detection tasks.
Nevertheless, studying the inductive biases in DMs that lead to this new pathological behaviour of likelihoods on generated data is an interesting open question requiring future research.

\clearpage
\section{A Mathematical Link between Probability Mass and Local Intrinsic Dimension} \label{appx:probability-mass-and-lid}

Here we make the association between probability mass and LID more mathematically concrete. This connection is enabled by a surprising result linking Gaussian convolutions and LID \citep{loaiza2022diagnosing, tempczyk2022lidl}. Intuitively, adding high-dimensional but low-variance Gaussian noise corrupts $p_\theta$ more easily when $p_\theta$ concentrates around low-dimensional regions (see Figure 1 from \citet{tempczyk2022lidl}). Comparing $p_\theta$ convolved with noise for different noise levels allows one to infer LID from the rate at which $p_\theta$ is corrupted as the noise increases. More formally, defining the convolution between a model density $p_\theta$ and a Gaussian with log standard deviation $r$ as 
\begin{align}\label{eq:rho_def}
    \rho_r(\vect{x}) \coloneqq [p_\theta(\cdot )* \mathcal{N}(\ \cdot \ ; \vect{0}, e^{2r}\mI_d)](\vect{x}) =\int p_\theta(\vect{x}{-}\vect{x}')\mathcal{N}(\vect{x}'; \vect{0}, e^{2r}\mI_d) \rd \vect{x}',
\end{align}
\citet{tempczyk2022lidl} showed that under mild regularity conditions, for sufficiently negative $r$ (i.e.\ low variance noise),
\begin{equation} \label{eq:LIDL-results}
     \log \rho_r(\vect{x}) = r (\LID_\theta(\vect{x}) - d) + \mathcal{O}(1).
\end{equation}
\autoref{eq:LIDL-results} suggests that, for sufficiently negative $r$, the rate of change of $\log \rho_r(\vect{x})$ with respect to $r$ can be used to estimate LID, since 
\begin{equation}
    \frac{\partial}{\partial r} \log \rho_r(\vect{x}) \approx \LID_\theta(\vect{x}) - d.
\end{equation}
We will now link the above quantity to the probability mass that $p_\theta$ assigns around $\vect{x}$. Let $B_R(\vect{x}) \coloneqq \{\vect{x}' \in \mathcal{X} : \Vert \vect{x}' - \vect{x} \Vert_2^2 \leq R^2\}$ be an $\ell_2$ ball of radius $R$ around $\vect{x}$. The probability that $p_\theta$ assigns to this ball is
\begin{align}\label{eq:proxy-formula}
        \mathbb{P}_\theta \left( \vect{x}' \in B_{R}(\vect{x}) \right) = \int_{B_{R}(\vect{x})} p_\theta(\vect{x}')  \rd \vect{x}' = \vol(B_{R}(\vect{0})) \cdot \left[p_\theta(\cdot) * \mathcal{U}(\, \cdot \, ;B_{R}(\vect{0}))\right] (\vect{x}),
\end{align}
where $\vol(B)$ denotes the $d$-dimensional Lebesgue measure of $B$ (i.e.\ its volume), and $\mathcal{U}(\, \cdot \, ;B)$ is the density of the uniform distribution on $B$ -- we note that $\vol(B_R(\vect{x}))$ is not strictly equivalent to the volume defined in \autoref{eq:vol}, as the latter depends on $p_\theta$ and ``ignores directions along which $p_\theta$ is negligible''. We now leverage the standard and well-known result that in high dimensions, the uniform distribution on the ball is approximately Gaussian, $\mathcal{U}(\, \cdot \, ;B_{e^r \sqrt{d}}(\vect{0})) \approx \mathcal{N}(\, \cdot \, ;\vect{0}, e^{2r}\mI_d)$.\footnote{Readers unfamiliar with this can see \citet{saremi2019neural} for a related derivation in a machine learning context. Note that this derivation shows Gaussians are approximately uniform on the boundary of the ball, but another classic result is that, in high dimensions, the majority of the mass of the ball lies near its boundary (see e.g.\ \citet{wegner2021lecture}), so that uniform distributions on the ball or its boundary are also approximately equal. Similarly, a textbook derivation shows that when the dimensionality $d$ is large, almost all the probability mass of a standard Gaussian is concentrated in an annulus at radius $\sqrt{d}$ \cite{blum2020foundations}.} This result combined with \autoref{eq:rho_def} suggests that, if we take $R=e^{r}\sqrt{d}$, we can approximate the log probability mass in \autoref{eq:proxy-formula} as
\begin{align}
    \log \mathbb{P}_\theta \left( \vect{x}' \in B_{e^{r}\sqrt{d}}(\vect{x}) \right) \approx \log \vol(B_{e^{r}\sqrt{d}}(\vect{0})) + \log \rho_r(\vect{x}).
\end{align}
Differentiating with respect to $r$ then yields
\begin{align}\label{eq:lid_mass_relation}
    \dfrac{\partial}{\partial r} \log \mathbb{P}_\theta \left( \vect{x}' \in B_{e^{r}\sqrt{d}}(\vect{x}) \right) \approx \LID_\theta(\vect{x}) + \left[ \dfrac{\partial}{\partial r}\log \vol(B_{e^{r}\sqrt{d}}(\vect{0})) - d \right],
\end{align}
which establishes a clear relationship between (log) probability mass and LID. Note that the second term on the right hand side of \autoref{eq:lid_mass_relation} is just a fixed function of $r$ which does not depend on the model $p_\theta$ nor on $\vect{x}$, so that thresholding this quantity is equivalent to thresholding $\LID_\theta(\vect{x})$ (whenever the model and $r$ are kept fixed). In other words, small/large values of $\LID_\theta(\vect{x})$ are equivalent to small/large values of the rate of change of the log probability mass around $\vect{x}$, i.e.\ $\tfrac{\partial}{\partial r} \log \mathbb{P}_\theta( \vect{x}' \in B_{e^{r}\sqrt{d}}(\vect{x}))$. In turn, our dual threshold algorithm described in \autoref{sec:all_together} can be interpreted as thresholding on an estimate of $\tfrac{\partial}{\partial r} \log \mathbb{P}_\theta( \vect{x}' \in B_{e^{r}\sqrt{d}}(\vect{x}))$ rather than on $\widehat{\LID}_\theta(\vect{x})$, which matches the understanding that we derived in \autoref{sec:method}: even if $\log p_\theta(\vect{x})$ is large, if $\log \mathbb{P}_\theta( \vect{x}' \in B_{e^{r}\sqrt{d}}(\vect{x}))$ increases slowly as a function of $r$ (i.e.\ $\LID_\theta(\vect{x})$ is small, this corresponds to case $(ii)$ in \autoref{sec:all_together}), we can sensibly expect $p_\theta$ to assign lower probability mass to a small ball around $\vect{x}$ as compared to the case where both $\log p_\theta(\vect{x})$ is large and $\log \mathbb{P}_\theta( \vect{x}' \in B_{e^{r}\sqrt{d}}(\vect{x}))$ increases quickly (which now corresponds to case $(iii)$ in \autoref{sec:all_together}). We can thus understand our dual threshold method as attempting to classify points as in-distribution when they have (relatively) large probability mass around them.

\clearpage
\section{LID Estimation and Setting the Threshold} \label{appx:lid-estimator-eval}

While setting a constant threshold $\tau$ can yield effective OOD detection results --- as demonstrated in \autoref{fig:fmnist_mnist_scatterplots} where we set it to an infinitesimal value of $\tau=10^{-10}$ --- choosing the right value of $\tau$ remains crucial for good performance across tasks. From a numerical perspective setting $\tau$ to an excessively small value would result in the LID estimator always predicting the ambient dimension, while setting it to an excessively large value will result in our estimator predicting $0$.
While \citet{horvat2022} and \citet{stanczuk2022your} offer thorough methods for setting the threshold $\tau$ and estimating intrinsic dimension, our focus is primarily on effective LID estimation for OOD detection. Consequently, we adopt a straightforward and rapid approach for LID estimation that behaves well for our intents and purposes. 

One sensible way of setting $\tau$ is to calibrate it based on another model-free estimator of LID using the training data. In particular, we perform local principal component analysis (LPCA) which is a model-free method for LID estimation. LPCA is similar to the LID estimator in \citet{horvat2022} which also uses the concept of local linearizations. We use the \texttt{scikit-dimension} \citep{bac2021} implementation and use the algorithm introduced by \citet{fukunaga1971algorithm} with \texttt{alphaFO} set to $0.001$ to estimate the average LID of our training data. Then $\tau$ is set so that $\textrm{LID}_\theta$ estimates of the training dataset match the LPCA average.

To increase efficiency, we select a random set of 80 data points from our training set as representative samples.  We then employ a binary search to fine-tune $\tau$. During each iteration of the binary search, we compare the average $\LID_\theta$ of our subsamples with the intrinsic dimension determined by LPCA. If the average $\LID_\theta$ is lower, we increase $\tau$; otherwise, we decrease it. We initially set $\tau$'s binary search range between $0$ and $10^{10}$, representing a wide range of plausible thresholds. Binary search is then executed in $50$ steps to accurately ascertain a value of $\tau$.
\autoref{tab:necessity_of_adaptive_scaling} represents three distinct scenarios to assess how to set $\tau$ optimally for NF models. In the first two rows, $\tau$ is held constant across datasets, while in the third, $\tau$ is dynamically adjusted to each dataset based on the above approach. Although there is a minor performance drop in the FMNIST vs.\ MNIST comparison, this is offset by a notable enhancement in the CIFAR10 vs.\ SVHN case. This significant improvement further justifies our preference for this method of obtaining the threshold rather than setting it to a fixed value as a hyperparameter.

\begin{table}[t]\captionsetup{font=footnotesize}
\centering
\caption{NF-based OOD detection performance (AUC-ROC) for various $\tau$ thresholds compared to using dataset-specific values of $\tau$ based on LPCA estimates of LID (higher is better).}
\label{tab:necessity_of_adaptive_scaling}
\begin{tabularx}{0.8\textwidth}{lcccc}
\toprule
 & \scriptsize{FMNIST vs. MNIST}& \scriptsize{MNIST vs. FMNIST} & \scriptsize{CIFAR10 vs. SVHN} & \scriptsize{SVHN vs. CIFAR10}\\
\midrule
$\tau=10^{-10}$ & $\mathbf{0.961}$ & $1.000$ & $0.730$ & $0.987$ \\
$\tau=4.5 \times 10^{-5}$ & $0.957$ & $1.000$ & $0.737$ & $0.987$ \\
Using LPCA & $0.951$ & $1.000$ & $\mathbf{0.936}$ & $0.987$ \\
\bottomrule
\end{tabularx}
\end{table}

\begin{figure}\captionsetup{font=footnotesize} 
  \begin{center}
    \begin{subfigure}[b]{0.48\linewidth} 
      \centering
      \includegraphics[scale=0.45]{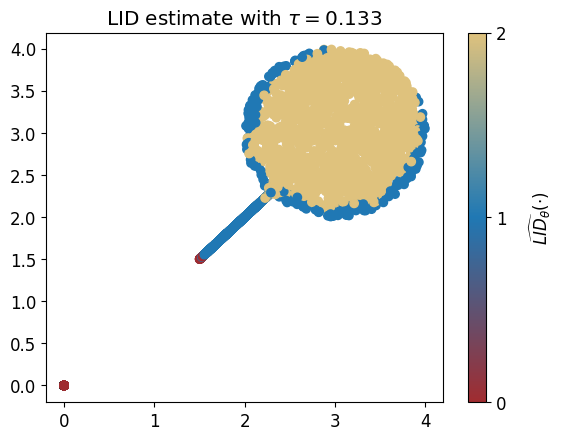}
      \caption{LID estimates for a DM.}
      \label{subfig:example_a_1}
    \end{subfigure}
    \hspace{0.01\linewidth} 
    \begin{subfigure}[b]{0.48\linewidth}
      \centering
      \includegraphics[scale=0.45]{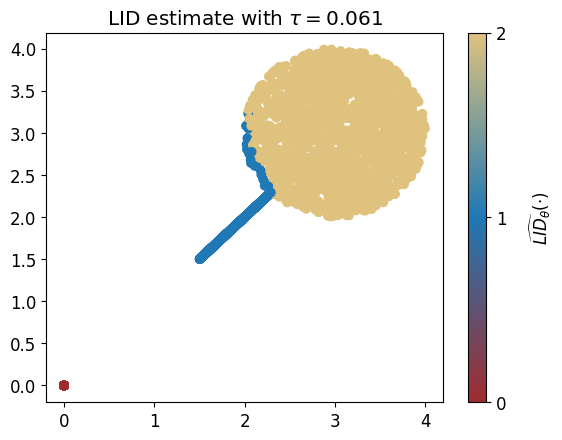}
      \caption{LID estimates for an NF.}
      \label{subfig:example_b_1}
    \end{subfigure}
  \end{center}
  \captionsetup{belowskip=-8pt}
  \caption{LID estimates using our LPCA approach to set thresholds. Colours indicate $\widehat{\LID_\theta}$ for a: \textbf{(a)} DM; \textbf{(b)} NF.}
  \label{fig:lollipop}
\end{figure}

To assess the LID estimates across points with varying dimensionalities, we utilize a 2D lollipop dataset \citep{tempczyk2022lidl} depicted in \autoref{fig:lollipop}. This dataset comprises points uniformly sampled from three distinct submanifolds: $(i)$ the ``candy'' portion with an intrinsic dimension of $2$; $(ii)$ the ``stick'' with an intrinsic dimension of $1$; and $(iii)$ an isolated point with an intrinsic dimension of $0$. Utilizing LPCA, which accurately estimates the average LID, our LPCA-based method correctly sets the threshold $\tau$. As shown in \autoref{fig:lollipop}, the LID estimates are generally precise on both NFs and DMs, with deviations only near the boundaries of the submanifolds, which is arguably appropriate behaviour.

\begin{figure}\captionsetup{font=footnotesize} 
  \begin{center}
    \begin{subfigure}[b]{0.48\linewidth} 
      \centering
      \includegraphics[scale=0.33]{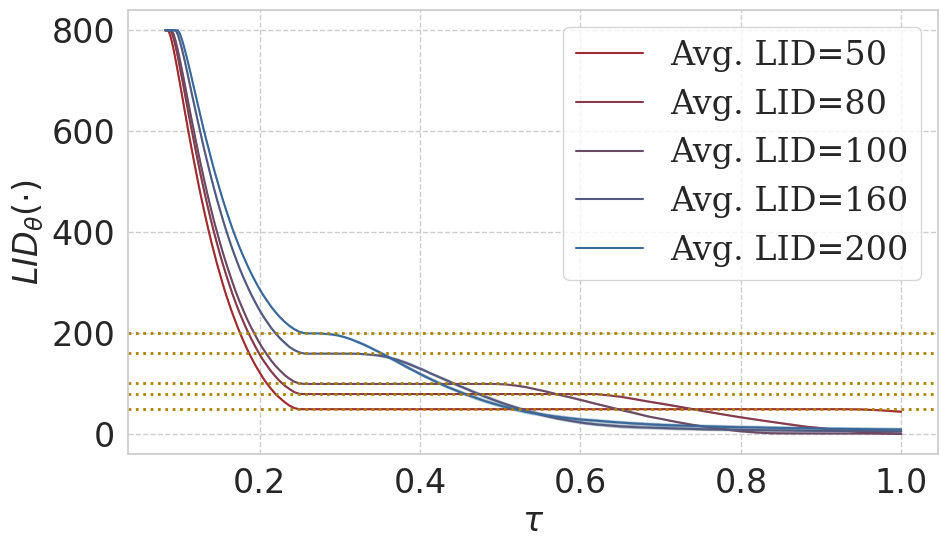}
      \caption{LID estimates for a DM.}
      \label{subfig:example_a_2}
    \end{subfigure}
    \hspace{0.01\linewidth} 
    \begin{subfigure}[b]{0.48\linewidth}
      \centering
      \includegraphics[scale=0.33]{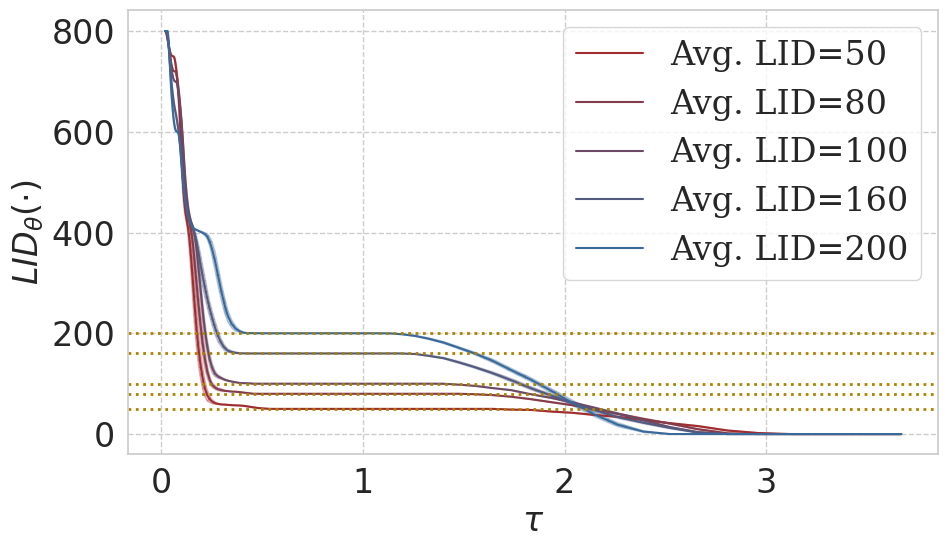}
      \caption{LID estimates for an NF.}
      \label{subfig:example_b_2}
    \end{subfigure}
  \end{center}
  \captionsetup{belowskip=-8pt}
  \caption{Average LID estimates for datapoints in $800$-dimensional datasets with low intrinsic dimension across various threshold values. The dotted lines correspond to the the LID estimate obtained from LPCA. Results are shown for: \textbf{(a)} DMs; \textbf{(b)} NFs.}
  \label{fig:high_dim_lid}
\end{figure}

In addition, to test our estimator in high-dimensional environments, we examine five distinct datasets, each with $800$ ambient dimensions but varying intrinsic dimensionalities of $50$, $80$, $100$, $160$, and $200$. For each dataset, we initially generate $10{,}000$ samples from an isotropic Gaussian in the respective intrinsic dimension and then replicate individual elements to expand the ambient dimensionality to $800$. We then train both NFs and DMs on these datasets, adjusting the threshold values $\tau$ to derive the results presented in \autoref{fig:high_dim_lid}. A key observation is that the model LID estimates stabilize at the actual intrinsic dimension for a range of thresholds $\tau$. 
Moreover, LPCA effectively aligns with this stabilization point, enabling our binary search method to precisely estimate the optimal threshold $\tau$ -- thus giving us confidence in our LID estimates.

\clearpage
\section{Experimental Details and Additional Experiments}\label{app:details}

\subsection{Hyperparameter Setting for Normalizing Flows} \label{appx:flow_hp_setting}
We trained both Glow \citep{kingma2018glow} and RQ-NSFs \citep{durkan2019neural} on our datasets, with the hyperparameters detailed in \autoref{tab:hyperparameters}. Specifically, while Glow utilized an affine coupling layer, we adopted RQ-NSF's piecewise rational quadratic coupling with two bins and linear tails capped at $1$. 
In \autoref{fig:grayscale_samples} and \autoref{fig:RGB_samples}, we highlight failure cases of the Glow architecture. The artifacts, particularly in CelebA, Tiny ImageNet, and Omniglot samples, stem from the affine coupling layers' unfavourable numerical properties. In contrast, the RQ-NSF architectures showed no such issues, leading us to adopt them for subsequent experiments.

\begin{table}[t]\captionsetup{font=footnotesize}
\centering
\caption{Essential hyperparameter settings for the normalizing flow models.}
\label{tab:hyperparameters}
\begin{tabularx}{0.9\textwidth}{p{3.9cm}>{\centering\arraybackslash}m{9.2cm}}
\toprule
Property & Model Configuration \\
\midrule
Learning rate & $1 \times 10^{-3}$ \\
Gradient Clipping & Value based (max = $1.0$)\\
Scheduler & \verb|ExponentialLR| (with a factor of $0.99$)\\
Optimizer & \verb|AdamW| \\
Weight decay & $5 \times 10^{-5}$ \\
Batch size & $128$ \\
Epochs & $400$ \\
\midrule
Transform blocks& Actnorm $\to$ $(1\times1)$ Convolution $\to$ Coupling\\
Number of multiscale levels & $7$ levels\\
Coupling layer backbone & ResNet (channel size $ = 64$, \# blocks = $2$, dropout = $0.2$)\\
Masking scheme & Checkerboard\\
Latent Space & Standard isotropic Gaussian\\
\midrule
Data pre-processing & Dequantization \& Logit scaling\\
Data shape & $28 \times 28 \times 1$ for grayscale and $32 \times 32 \times 3$ for RGB\\
\bottomrule
\end{tabularx}
\end{table}

\begin{figure}[b]\captionsetup{font=footnotesize}
    \centering
    
    \begin{subfigure}{0.24\textwidth}
        \includegraphics[width=\linewidth]{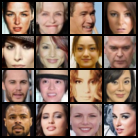} 
        \caption{Original samples from the test split. \\ \, }
    \end{subfigure}%
    \hfill
    \begin{subfigure}{0.24\textwidth}
        \includegraphics[width=\linewidth]{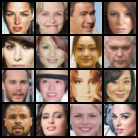}
        \caption{Reconstructed samples from the test split.}
    \end{subfigure}%
    \hfill
    \begin{subfigure}{0.24\textwidth}
        \includegraphics[width=\linewidth]{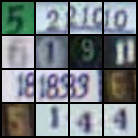}
        \caption{Original samples from the OOD dataset.}
    \end{subfigure}%
    \hfill
    \begin{subfigure}{0.24\textwidth}
        \includegraphics[width=\linewidth]{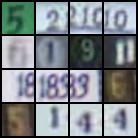}
        \caption{Reconstructed samples from the OOD dataset.}
    \end{subfigure}
    
    \caption{Numerical invertibility: \textbf{(a-b)} A random batch of samples and their reconstructions from the test split of an RQ-NSF model trained on CelebA. \textbf{(c-d)} A random batch of samples and their reconstructions from the OOD dataset, SVHN.}
    \label{fig:numerical_invertibility}
\end{figure}

\begin{figure}[t]\captionsetup{font=footnotesize}
    \centering
    
    \begin{subfigure}{0.24\textwidth}
        \includegraphics[width=\linewidth]{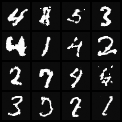} 
        \caption{Glow model trained on MNIST.}
    \end{subfigure}%
    \hfill
    \begin{subfigure}{0.24\textwidth}
        \includegraphics[width=\linewidth]{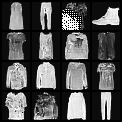}
        \caption{Glow model trained on FMNIST.}
    \end{subfigure}%
    \hfill
    \begin{subfigure}{0.24\textwidth}
        \includegraphics[width=\linewidth]{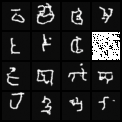}
        \caption{Glow model trained on Omniglot.}
    \end{subfigure}%
    \hfill
    \begin{subfigure}{0.24\textwidth}
        \includegraphics[width=\linewidth]{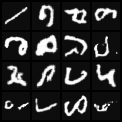}
        \caption{Glow model trained on EMNIST.}
    \end{subfigure}

    \vspace{1em} 
    
    \begin{subfigure}{0.24\textwidth}
        \includegraphics[width=\linewidth]{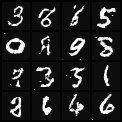}
        \caption{RQ-NSF model trained on MNIST.}
    \end{subfigure}%
    \hfill
    \begin{subfigure}{0.24\textwidth}
        \includegraphics[width=\linewidth]{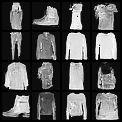}
        \caption{RQ-NSF model trained on FMNIST.}
    \end{subfigure}%
    \hfill
    \begin{subfigure}{0.24\textwidth}
        \includegraphics[width=\linewidth]{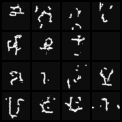}
        \caption{RQ-NSF model trained on Omniglot.}
    \end{subfigure}%
    \hfill
    \begin{subfigure}{0.24\textwidth}
        \includegraphics[width=\linewidth]{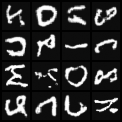}
        \caption{RQ-NSF model trained on EMNIST.}
    \end{subfigure}
    
    \caption{Samples generated from models trained on the grayscale collection: due to numerical properties of affine coupling layers, Glow models tend to produce artifacts in their generated data.}
    \label{fig:grayscale_samples}
\end{figure}

\begin{figure}[!htb]\captionsetup{font=footnotesize}
    \centering
    
    \begin{subfigure}{0.24\textwidth}
        \includegraphics[width=\linewidth]{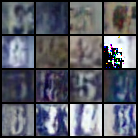} 
        \caption{Glow model trained on ~SVHN.}
    \end{subfigure}%
    \hfill
    \begin{subfigure}{0.24\textwidth}
        \includegraphics[width=\linewidth]{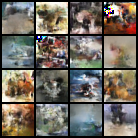}
        \caption{Glow model trained on CIFAR10.}
    \end{subfigure}%
    \hfill
    \begin{subfigure}{0.24\textwidth}
        \includegraphics[width=\linewidth]{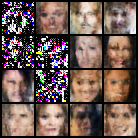}
        \caption{Glow model trained on CelebA.}
    \end{subfigure}%
    \hfill
    \begin{subfigure}{0.24\textwidth}
        \includegraphics[width=\linewidth]{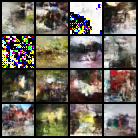}
        \caption{Glow model trained on Tiny ImageNet.}
    \end{subfigure}

    \vspace{1em} 
    
    \begin{subfigure}{0.24\textwidth}
        \includegraphics[width=\linewidth]{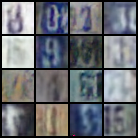}
        \caption{RQ-NSF model trained on SVHN. \\ \,}
    \end{subfigure}%
    \hfill
    \begin{subfigure}{0.24\textwidth}
        \includegraphics[width=\linewidth]{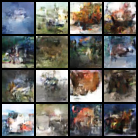}
        \caption{RQ-NSF model trained on CIFAR10. \\ \,}
    \end{subfigure}%
    \hfill
    \begin{subfigure}{0.24\textwidth}
        \includegraphics[width=\linewidth]{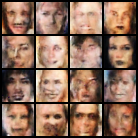}
        \caption{RQ-NSF model trained on CelebA. \\ \,}
    \end{subfigure}%
    \hfill
    \begin{subfigure}{0.24\textwidth}
        \includegraphics[width=\linewidth]{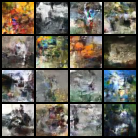}
        \caption{RQ-NSF model trained on Tiny ImageNet.}
    \end{subfigure}
    
    \caption{Samples generated from models trained on the RGB collection: the artifacts in Glow models are apparent.}
    \label{fig:RGB_samples}
\end{figure}

In the context of OOD detection, expressive architectures sometimes face issues of numerical non-invertibility and exploding inverses, particularly with OOD samples \citep{behrmann2021understanding}.  While expressive NFs adeptly fit data manifolds, their mapping from a full-dimensional space to a lower-dimensional one can cause non-invertibility, especially in OOD datapoints. \citet{behrmann2021understanding} specifically identified non-invertibility examples in Glow models on OOD data. Contrarily, the RQ-NSFs we trained according to the hyperparameter setup in \autoref{tab:hyperparameters} demonstrated full reconstruction on OOD data, as depicted in \autoref{fig:numerical_invertibility}. This is another reason why we chose RQ-NSFs.

We used an NVIDIA Tesla V100 SXM2 with $7$ hours of GPU time to train each of the models.

\subsection{Hyperparameter Setting for Diffusion Models} \label{appx:diffusion_hp_setting}

\begin{table}[t]\captionsetup{font=footnotesize}
\centering
\caption{Essential hyperparameter settings for the Diffusion models.}
\label{tab:hyperparameters-diffusions}
\begin{tabularx}{0.9\textwidth}{p{3.9cm}>{\centering\arraybackslash}m{9.2cm}}
\toprule
Property & Model Configuration \\
\midrule
Learning rate & $5 \times 10^{-5}$ \\
Gradient Clipping & Value based (max = $1.0$)\\
Optimizer & \verb|Adam| \\
Batch size & $128$ \\
Epochs & $200$ \\
\midrule
Score-matching weighting & Likelihood weighting $\lambda(t) \coloneqq \Var(\vect{x}_{T-t} \mid \vect{x}_0)$\\
SDE dynamics & Variance preserving with $\beta(t) \coloneqq 0.1 + 20t$\\
Maximum time & $T = 1$\\
\midrule
UNet \# channels& $(2 \times 128) \to (2 \times 256) \to (2 \times 512)$\\
Attention & The penultimate UNet block performs spatial self-attention\\
Down/Up-sampling Blocks & ResNet\\
\midrule
Data pre-processing & Dequantization \& Scaling pixels between $[0, 1]$\\
Data shape & Resize everything to $32 \times 32$\\
\bottomrule
\end{tabularx}
\end{table}

\begin{figure}[H]\captionsetup{font=footnotesize}
    \centering
    
    \begin{subfigure}{0.24\textwidth}
        \includegraphics[width=\linewidth]{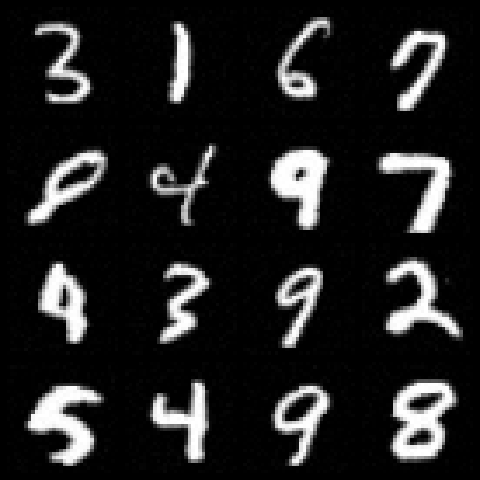} 
        \caption{DM samples on MNIST.}
    \end{subfigure}%
    \hfill
    \begin{subfigure}{0.24\textwidth}
        \includegraphics[width=\linewidth]{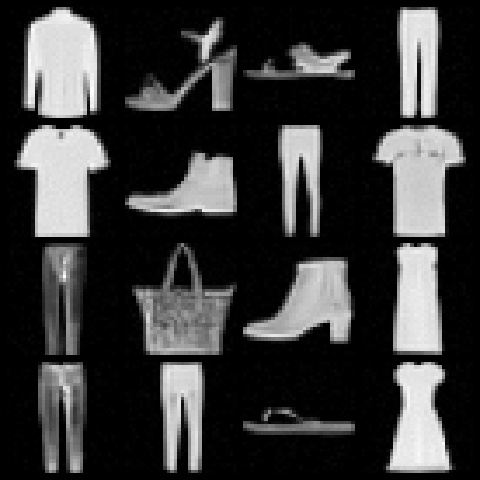}
        \caption{DM samples on FMNIST.}
    \end{subfigure}%
    \hfill
    \begin{subfigure}{0.24\textwidth}
        \includegraphics[width=\linewidth]{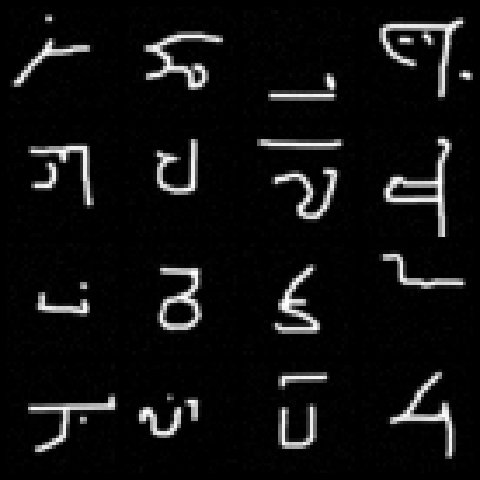}
        \caption{DM samples on Omniglot.}
    \end{subfigure}%
    \hfill
    \begin{subfigure}{0.24\textwidth}
        \includegraphics[width=\linewidth]{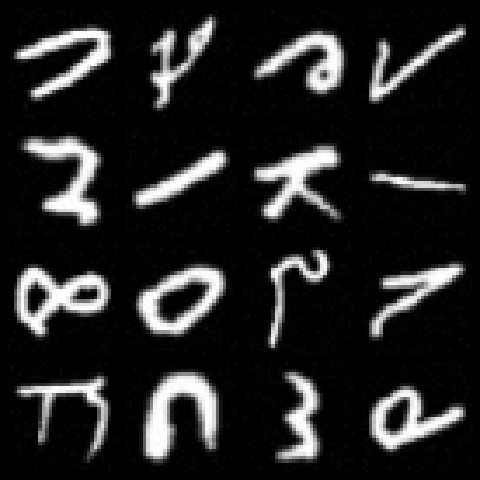}
        \caption{DM samples on EMNIST.}
    \end{subfigure}

    \vspace{1em} 
    
    \begin{subfigure}{0.24\textwidth}
        \includegraphics[width=\linewidth]{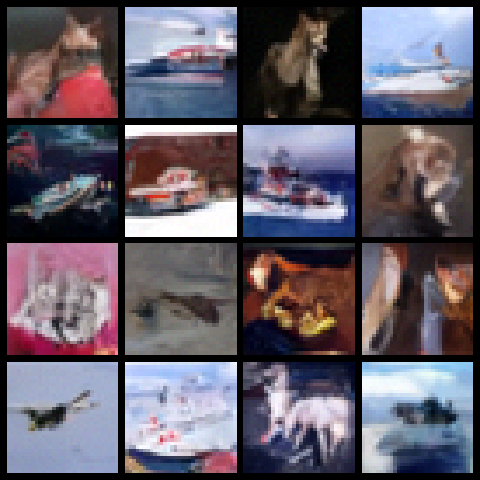}
        \caption{DM samples on CIFAR10.}
    \end{subfigure}%
    \hfill
    \begin{subfigure}{0.24\textwidth}
        \includegraphics[width=\linewidth]{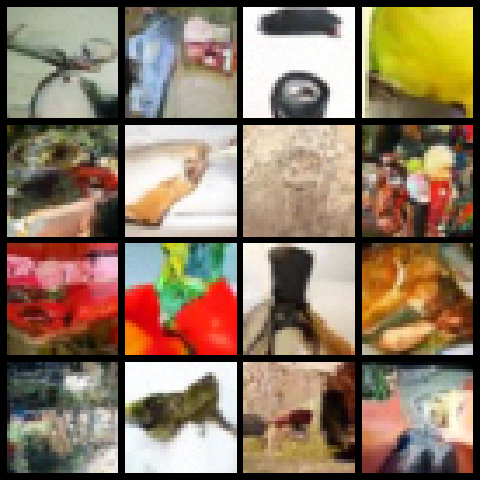}
        \caption{DM samples on Tiny ImageNet.}
    \end{subfigure}%
    \hfill
    \begin{subfigure}{0.24\textwidth}
        \includegraphics[width=\linewidth]{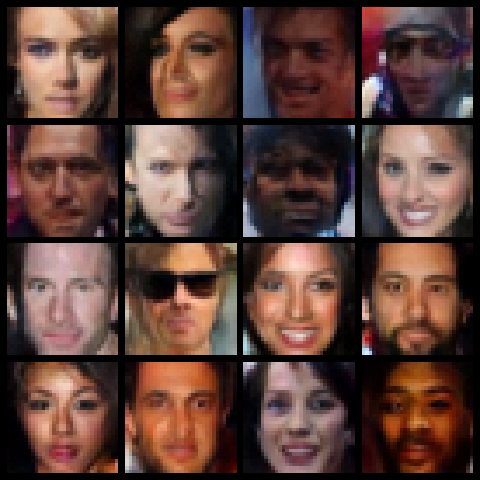}
        \caption{DM samples on CelebA.}
    \end{subfigure}%
    \hfill
    \begin{subfigure}{0.24\textwidth}
        \includegraphics[width=\linewidth]{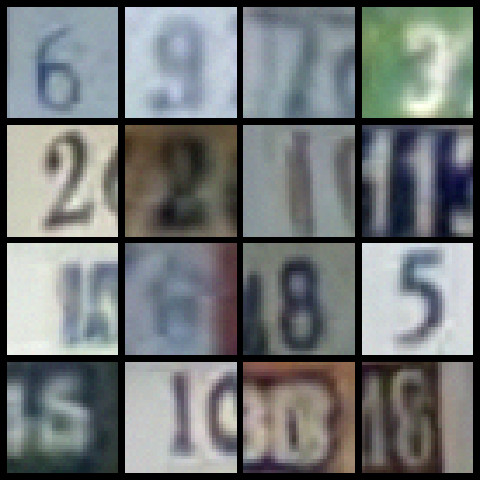}
        \caption{DM samples on SVHN.}
    \end{subfigure}
    
    \caption{Samples generated from DMs trained on all the datasets: the number of diffusion steps to generate these samples are $1000$.}
    \label{fig:DM_samples}
\end{figure}

We utilized UNet backbones from the \verb|diffusers| library \citep{von-platen-etal-2022-diffusers} for training our diffusion models. While the library provided the foundational architecture, we manually implemented additional functionalities such as LID estimation and log likelihood computations. The key hyperparameters guiding our training process are detailed in \autoref{tab:hyperparameters-diffusions}. Samples generated by these models, as illustrated in \autoref{fig:DM_samples}, demonstrate a markedly superior quality compared to those produced by NFs. We used an NVIDIA Tesla V100 SXM2 with on average $4$ hours of GPU time to train each of the models.

\begin{figure}\captionsetup{font=footnotesize}
    \centering
    
    \begin{subfigure}{0.48\textwidth}
        \includegraphics[width=\linewidth]{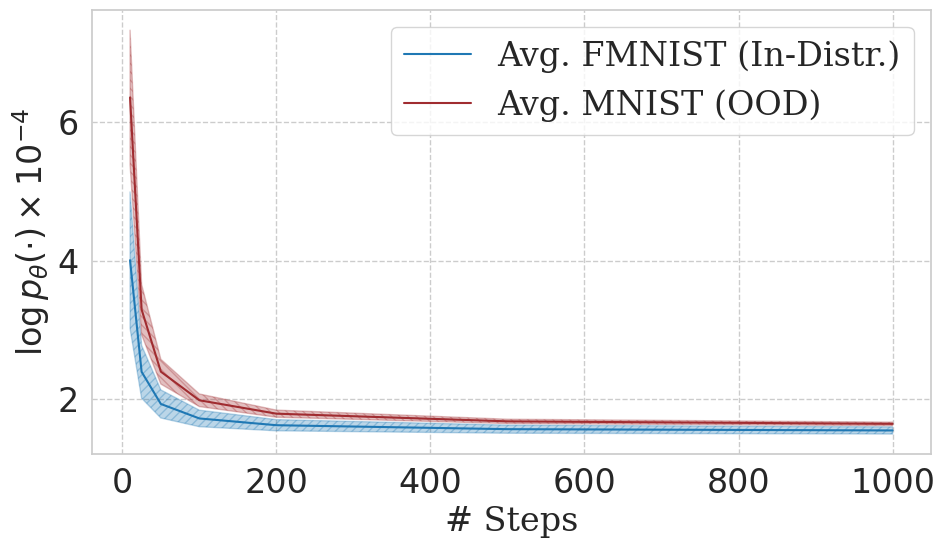} 
        \caption{Likelihoods calculated for a model trained on FMNIST and evaluated on both FMNIST (blue) and MNIST (red); the numbers of Euler iterations are varied across the x-axis and the log probabilities mean and variance are shown on the y-axis.}
        \label{subfig:fmnist_vs_mnist_hp_tune}
    \end{subfigure}%
    \hfill
    \begin{subfigure}{0.48\textwidth}
        \includegraphics[width=\linewidth]{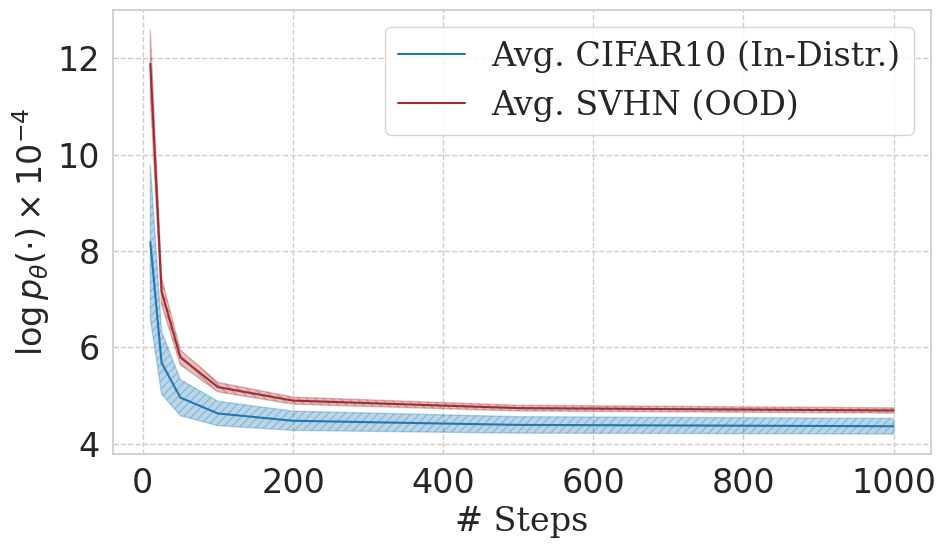}
        \caption{Likelihoods calculated for a model trained on CIFAR10 and evaluated on both CIFAR10 (blue) and SVHN (red); the numbers of Euler iterations are varied across the x-axis and the log probabilities mean and variance are shown on the y-axis.}
        \label{subfig:cifar10_vs_svhn_hp_tune}
    \end{subfigure}%

    \vspace{1em} 
    
    \begin{subfigure}{0.48\textwidth}
        \includegraphics[width=\linewidth]{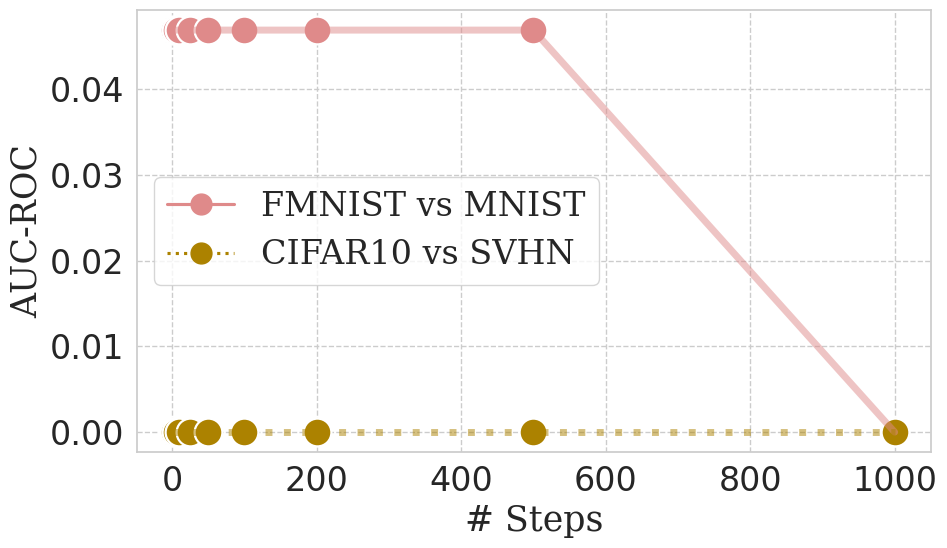} 
        \caption{AUC-ROC evaluated by a single threshold OOD detector on likelihoods for two pathological tasks; the number of Euler iterations are varied across the x-axis and the AUC-ROC is shown on the y-axis.}
        \label{subfig:num_steps_sweep}
    \end{subfigure}%
    \hfill
    \begin{subfigure}{0.48\textwidth}
        \includegraphics[width=\linewidth]{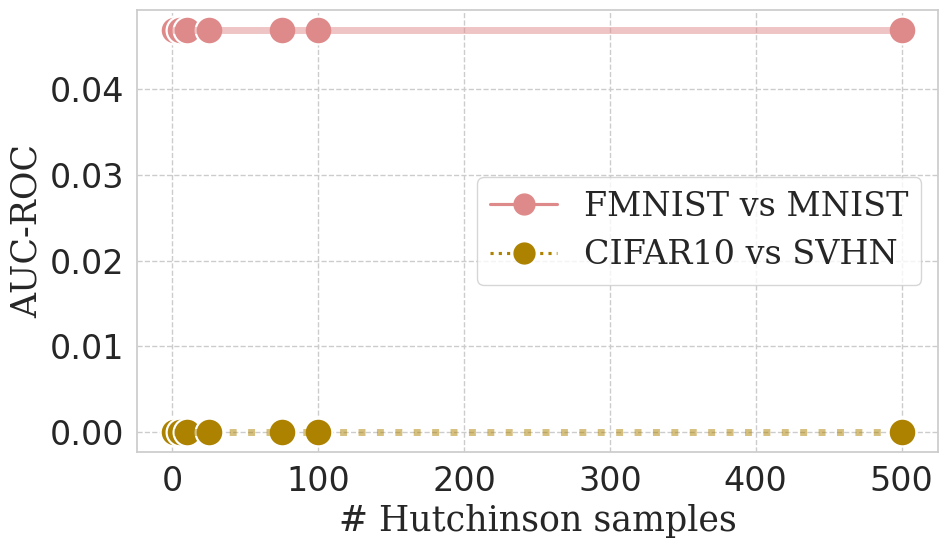}
        \caption{AUC-ROC evaluated by a single threshold OOD detector on likelihoods for two pathological tasks; the number of samples for Hutchinson trace estimation is varied across the x-axis and the AUC-ROC is shown on the y-axis.}
        \label{subfig:num_samples_sweep}
    \end{subfigure}%
    
    \caption{The pathological behaviour holds even for strong hyperparameter settings for likelihood evaluation: \textbf{(a-b)} increasing the number of Euler steps produces more accurate likelihood estimates but does not change the ordering of the in-distribution and OOD estimates; \textbf{(c-d)} increasing the number of Euler steps and trace estimation samples does not improve the performance of a likelihood-based OOD detector.}
    \label{fig:DM_pathology}
\end{figure}

\paragraph{Computing Likelihoods} To calculate the likelihood, one needs to solve an appropriate ordinary differential equation (for more details, please refer to \citet{song2020score}). We use a standard Euler numerical solver and take $25$ iterations to compute the likelihood estimates. Moreover, at each step of the solver, one needs to compute the trace of the score's Jacobian utilizing the Hutchinson trace estimator \citep{hutchinson1989stochastic}. We set the number of samples for trace estimation to $25$. These hyperparameters are chosen to ensure tractability of estimating likelihoods across entire datasets of size $2^{15}$. 
In addition to that, we performed an extensive hyperparameter sweep to make sure that the pathology occurs even when likelihoods are computed more accurately with a stronger hyperparameter setting --- i.e. with more steps in the differential equation solver or with more Hutchinson samples to estimate the trace.
For further evaluation, we have picked two pathological OOD detection scenarios: FMNIST vs.\ MNIST and CIFAR10 vs.\ SVHN where likelihoods inflate on OOD data.
In \autoref{subfig:fmnist_vs_mnist_hp_tune} and \autoref{subfig:cifar10_vs_svhn_hp_tune} we hold the number of Hutchinson samples at a constant $500$ and vary the number of steps in the Euler solver. As the number of steps increases, the likelihood estimate concentrates more accurately around the true likelihood value. However, even with a larger number of steps, the ordering of OOD likelihoods versus their in-distribution counterparts does not change.
We performed a similar experiment where we held the number of steps at a constant $500$ and varied the numbers of Hutchinson samples; we observed no substantial change in the mean and standard deviation of the likelihoods beyond $25$ samples. 
Finally, we also compute AUC-ROC for single threshold classifiers on likelihood values for these tasks to quantify the pathology.
In \autoref{subfig:num_samples_sweep}, we maintain a constant step count of $25$ while varying the number of Hutchinson samples. This adjustment reveals negligible variations in the AUC metric across both evaluated tasks. Similarly, \autoref{subfig:num_steps_sweep} documents our experiment where we fix the Hutchinson samples at $25$ and modify the step count. Here, not only do we observe minimal fluctuations in the AUC-ROC metric with increasing steps, but we also note a deterioration in performance for the FMNIST vs.\ MNIST, indicating the pathology even becomes worse as the configuration for likelihoods becomes more accurate. These results rule out the possibility that the pathological behaviour of likelihoods for OOD detection in DMs was caused by poor density evaluation.

\subsection{Ablations on Variance Exploding Diffusion Models} \label{appx:vesde_vs_vpsde}

\citet{stanczuk2022your} propose their LID estimation technique for variance exploding DMs, we trained a series of variance exploding DMs with the same hyperparameter setting as \autoref{tab:hyperparameters-diffusions} but with an appropriate noise scheduler and assessed their performance in OOD detection, as detailed in \autoref{tab:diffusion-ablation}. The results demonstrate that combining variance preserving LID with likelihoods yields more consistent outcomes for OOD detection, leading us to favour this type of DM. 
We hypothesize this is due to either one (or a combination) of the following reasons that render the likelihood estimates inaccurate: 
$(i)$ Variance preserving DMs are designed in such a way that $p_T$ is very close to an isotropic Gaussian. In practice $\hat{p}_T$ is thus chosen as an isotropic Gaussian for these models. On the other hand, for variance exploding DMs, $p_T$ does not converge, and $\hat{p}_T$ is simply set to a Gaussian with large variance. We thus hypothesize that the greater mismatch between $p_T$ and $\hat{p}_T$ in the variance exploding setting might in turn render likelihoods less reliable, thereby adversely affecting the performance of OOD detection. 
$(ii)$ The input to the score networks in variance exploding DMs is heterogeneous, meaning that at time $t$ the input $\vect{x}_t$ to the score network $s_\theta(\vect{x}_t, t)$ might be either extremely large or small in scale. In particular, for $t \gg 0$, $\vect{x}_t$ would be extremely noisy with large variance, and as $t \to 0^+$, $\vect{x}_t$ would be on the scale of the image data. This results in an unstable score network that can negatively impact the likelihood estimates. 
It is important to highlight that \citet{song2019generative} employ a technique named \texttt{CondInstanceNorm++} to address this challenge. Unfortunately, this method has not been incorporated into the existing \texttt{diffusers} diffusion architectures that we have used. Despite this omission, our variance preserving DMs yield satisfactory outcomes without necessitating such additional complexities.

\begin{table}\captionsetup{font=footnotesize}
\centering
\footnotesize
\caption{AUC-ROC (higher is better) at A (vs.) B tasks. Ablation between variance preserving and exploding variations of DMs.}
\begin{tabularx}{\textwidth}{l*{8}{Y}}
\toprule
~~~Trained on & \multicolumn{2}{c}{MNIST} & \multicolumn{2}{c}{FMNIST} & \multicolumn{2}{c}{CIFAR10} & \multicolumn{2}{c}{SVHN}\\
\cmidrule(r){2-3} \cmidrule(lr){4-5} \cmidrule(lr){6-7} \cmidrule(l){8-9}
OOD Dataset& {\scriptsize FMNIST} & {\scriptsize Omniglot} & {\scriptsize MNIST} & {\scriptsize Omniglot} & {\scriptsize SVHN} & {\scriptsize CelebA} & {\scriptsize CIFAR10} & {\scriptsize CelebA} \\
\midrule
Variance Exploding DM Likelihood & $\mathbf{0.999}$ & $\mathbf{1.000}$ & $0.211$ & $0.953$ & $0.088$ & $0.485$ & $\mathbf{0.990}$ & $\mathbf{0.998}$ \\
Variance Exploding DM Dual Threshold& $\mathbf{0.999}$ & $\mathbf{1.000}$ & $0.899$ & $0.954$ & $0.855$ & $0.485$ & $\mathbf{0.990}$ & $\mathbf{0.998}$ \\
Variance Preserving DM Likelihood & $0.996$ & $1.000$ & $0.240$ & $0.952$ & $0.064$ & $0.360$ & $0.996$ & $0.996$ \\
Variance Preserving DM Dual Threshold & $0.996$ & $1.000$ & $\mathbf{0.912}$ & $\mathbf{0.959}$ & $\mathbf{0.944}$ & $\mathbf{0.648}$ & $0.996$ & $0.996$ \\
\bottomrule
\end{tabularx}
\label{tab:diffusion-ablation}
\end{table}

\subsection{Extra Dataset Pairs and Results on the Generated Samples Pathologies} \label{appx:all_aucs_gen_pathologies}
\autoref{tab:boosting_AUC_NF} and \autoref{tab:boosting_AUC_DM} compare na\"ively detecting OOD based on likelihoods against our dual threshold method across all tasks for NFs and DMs, respectively. 
In these tables, we use the suffixes ``-train'', ``-test'', and ``-gen'' when talking about datasets to specify if we are referring to the train set, test set, or generated samples, respectively.
For datasets $A$ and $B$, ``$A$ vs.\ $B$'' indicates the OOD detection task that aims to distinguish $A$-test from $B$-test using the model $p_\theta$ pre-trained on $A$-train. 
When we write ``$A$-gen vs.\ $B$'', $A$-test is replaced by $A$-gen, but $p_\theta$ is still pre-trained on $A$-train.
Note that even for $A$-gen vs.\ $B$ tasks we calibrate $\tau$ using the training dataset.
For fairness across all tasks, for any dataset, a random set of $2^{15}=32768$ sampled datapoints with replacement has been considered. We can see an extremely consistent improvement, highlighting the relevance of dual thresholding. 

The tasks of the form $A$-gen vs.\ $B$ are especially relevant in our study as there is no discrepancy between the in-distribution and model manifolds by design, and thus any practical concerns about poor model fit and misalignment between generated and in-distribution samples would be addressed; this in turn tests our hypothesis in a more isolated manner.
Furthermore, in \autoref{tab:results-table-gen} we have benchmarked our method against the baselines of \autoref{tab:results-table} but replaced tasks $A$ vs.\ $B$ with $A$-gen vs.\ $B$. For all the $A$-gen vs.\ $B$ tasks, we use the same $\tau$ as we used for the corresponding $A$ vs.\ $B$ task, i.e.\ we do not recalibrate $\tau$. Overall, our dual threshold method outperforms \textit{all} the baselines on NFs, but falls short in some comparisons against DM baselines. We hypothesize this discrepancy is because the LID estimates from \citet{stanczuk2022your} are not sufficiently accurate.

\begin{table}[h!]
\centering
\footnotesize
\caption{AUC-ROC Results for NF Experiments: Comparing likelihood-only and dual-threshold methods (higher is better). The table is split into greyscale tasks (top) and RGB tasks (bottom). Entries showing over $10\%$ improvement when comparing dual thresholding to the likelihood-only counterpart are boldfaced.
}
    \begin{tabularx}{0.75\textwidth}{l*{4}{Y}}
    \toprule
    OOD Task Type & \multicolumn{2}{c}{$A$-gen vs.\ $B$}& \multicolumn{2}{c}{$A$ vs.\ $B$}\\
    \cmidrule(r){2-3} \cmidrule(lr){4-5}
    \leftcell{Dataset Pair \\$A$ and $B$}&  \makecell{{\scriptsize (AUC-ROC)} \\ Likelihood} &  \makecell{{\scriptsize (AUC-ROC)} \\ Dual Threshold}&  \makecell{{\scriptsize (AUC-ROC)} \\ Likelihood} &  \makecell{{\scriptsize (AUC-ROC)} \\ Dual Threshold}  \\
    \midrule
    {\scriptsize FMNIST and MNIST} & $0.000$ & $\mathbf{0.999}$ & $0.073$ & $\mathbf{0.951}$ \\ 
{\scriptsize FMNIST and EMNIST} & $0.001$ & $\mathbf{0.956}$ & $0.394$ & $\mathbf{0.596}$ \\ 
{\scriptsize FMNIST and Omniglot} & $0.000$ & $\mathbf{0.996}$ & $0.085$ & $\mathbf{0.864}$ \\ 
{\scriptsize Omniglot and FMNIST} & $0.153$ & $\mathbf{1.000}$ & $1.000$ & $1.000$ \\ 
{\scriptsize Omniglot and MNIST} & $0.016$ & $\mathbf{1.000}$ & $1.000$ & $1.000$ \\ 
{\scriptsize Omniglot and EMNIST} & $0.000$ & $\mathbf{1.000}$ & $0.984$ & $0.984$ \\ 
{\scriptsize EMNIST and FMNIST} & $0.038$ & $\mathbf{1.000}$ & $0.998$ & $0.998$ \\ 
{\scriptsize EMNIST and MNIST} & $0.000$ & $\mathbf{1.000}$ & $0.540$ & $\mathbf{0.806}$ \\ 
{\scriptsize EMNIST and Omniglot} & $0.000$ & $\mathbf{1.000}$ & $0.389$ & $\mathbf{0.824}$ \\ 
{\scriptsize MNIST and FMNIST} & $0.007$ & $\mathbf{0.999}$ & $1.000$ & $1.000$ \\ 
{\scriptsize MNIST and EMNIST} & $0.000$ & $\mathbf{1.000}$ & $0.987$ & $0.988$ \\ 
{\scriptsize MNIST and Omniglot} & $0.000$ & $\mathbf{1.000}$ & $0.796$ & $0.855$ \\ 
\midrule 

{\scriptsize Tiny and CelebA} & $0.639$ & $0.660$ & $0.821$ & $0.821$ \\ 
{\scriptsize Tiny and SVHN} & $0.030$ & $\mathbf{0.961}$ & $0.154$ & $\mathbf{0.913}$ \\ 
{\scriptsize Tiny and CIFAR100} & $0.694$ & $\mathbf{0.799}$ & $0.805$ & $0.831$ \\ 
{\scriptsize Tiny and CIFAR10} & $0.694$ & $\mathbf{0.787}$ & $0.805$ & $0.831$ \\ 
{\scriptsize CelebA and Tiny} & $0.938$ & $0.960$ & $0.906$ & $0.928$ \\ 
{\scriptsize CelebA and SVHN} & $0.148$ & $\mathbf{0.935}$ & $0.146$ & $\mathbf{0.949}$ \\ 
{\scriptsize CelebA and CIFAR100} & $0.945$ & $0.968$ & $0.921$ & $0.942$ \\ 
{\scriptsize CelebA and CIFAR10} & $0.948$ & $0.967$ & $0.921$ & $0.939$ \\ 
{\scriptsize SVHN and Tiny} & $0.972$ & $0.972$ & $0.989$ & $0.989$ \\ 
{\scriptsize SVHN and CelebA} & $0.984$ & $0.984$ & $0.996$ & $0.996$ \\ 
{\scriptsize SVHN and CIFAR100} & $0.967$ & $0.967$ & $0.986$ & $0.986$ \\ 
{\scriptsize SVHN and CIFAR10} & $0.970$ & $0.970$ & $0.987$ & $0.987$ \\ 
{\scriptsize CIFAR100 and Tiny} & $0.386$ & $\mathbf{0.448}$ & $0.477$ & $0.479$ \\  
{\scriptsize CIFAR100 and CelebA} & $0.226$ & $\mathbf{0.646}$ & $0.370$ & $\mathbf{0.638}$ \\ 
{\scriptsize CIFAR100 and SVHN} & $0.014$ & $\mathbf{0.941}$ & $0.072$ & $\mathbf{0.933}$ \\ 
{\scriptsize CIFAR100 and CIFAR10} & $0.403$ & $\mathbf{0.486}$ & $0.490$ & $0.491$ \\ 
{\scriptsize CIFAR10 and Tiny} & $0.376$ & $\mathbf{0.535}$ & $0.485$ & $0.491$ \\  
{\scriptsize CIFAR10 and CelebA} & $0.239$ & $\mathbf{0.724}$ & $0.391$ & $\mathbf{0.655}$ \\ 
{\scriptsize CIFAR10 and SVHN} & $0.014$ & $\mathbf{0.950}$ & $0.063$ & $\mathbf{0.936}$ \\ 
{\scriptsize CIFAR10 and CIFAR100} & $0.426$ & $\mathbf{0.602}$ & $0.521$ & $0.562$ \\ 
    \bottomrule
    \end{tabularx}
\label{tab:boosting_AUC_NF}
\end{table}

\begin{table}[h!]
\centering
\footnotesize
\caption{AUC-ROC Results for DM Experiments: Comparing likelihood-only and dual-threshold methods (higher is better). The table is split into greyscale tasks (top) and RGB tasks (bottom). Entries showing over $10\%$ improvement when comparing dual thresholding to the likelihood-only counterpart are boldfaced.
}
    \begin{tabularx}{0.75\textwidth}{l*{4}{Y}}
    \toprule
    OOD Task Type & \multicolumn{2}{c}{$A$-gen vs.\ $B$} & \multicolumn{2}{c}{$A$ vs.\ $B$} \\
    \cmidrule(r){2-3} \cmidrule(lr){4-5}
    \leftcell{Dataset Pair \\$A$ and $B$}&  \makecell{{\scriptsize (AUC-ROC)} \\ Likelihood} &  \makecell{{\scriptsize (AUC-ROC)} \\ Dual Threshold}&  \makecell{{\scriptsize (AUC-ROC)} \\ Likelihood} &  \makecell{{\scriptsize (AUC-ROC)} \\ Dual Threshold}  \\
    \midrule
    {\scriptsize MNIST and EMNIST} & $0.000$ & $\mathbf{1.000}$ &$0.846$ & $0.846$ \\ 
{\scriptsize MNIST and FMNIST} & $0.000$ & $\mathbf{0.999}$ &$0.996$ & $0.996$ \\ 
{\scriptsize MNIST and Omniglot} & $0.000$ & $\mathbf{0.980}$ &$1.000$ & $1.000$ \\ 
{\scriptsize EMNIST and MNIST} & $0.000$ & $\mathbf{1.000}$ &$0.830$ & $0.830$ \\ 
{\scriptsize EMNIST and FMNIST} & $0.000$ & $\mathbf{0.991}$ &$0.999$ & $0.999$ \\ 
{\scriptsize EMNIST and Omniglot} & $0.000$ & $\mathbf{1.000}$ &$1.000$ & $1.000$ \\ 
{\scriptsize FMNIST and MNIST} & $0.000$ & $\mathbf{1.000}$ &$0.240$ & $\mathbf{0.912}$ \\ 
{\scriptsize FMNIST and EMNIST} & $0.000$ & $\mathbf{1.000}$ &$0.339$ & $\mathbf{0.568}$ \\ 
{\scriptsize FMNIST and Omniglot} & $0.000$ & $\mathbf{1.000}$ &$0.952$ & $0.959$ \\ 
{\scriptsize Omniglot and MNIST} & $0.000$ & $\mathbf{0.971}$ &$0.995$ & $0.995$ \\ 
{\scriptsize Omniglot and EMNIST} & $0.000$ & $\mathbf{0.952}$ &$1.000$ & $1.000$ \\ 
{\scriptsize Omniglot and FMNIST} & $0.000$ & $\mathbf{0.979}$ &$1.000$ & $1.000$ \\ 
\midrule 

{\scriptsize SVHN and Tiny} & $0.768$ & $\mathbf{0.891}$ &$0.996$ & $0.996$ \\ 
{\scriptsize SVHN and CIFAR10} & $0.774$ & $\mathbf{0.833}$ &$0.996$ & $0.996$ \\ 
{\scriptsize SVHN and CelebA} & $0.608$ & $\mathbf{0.802}$ &$0.996$ & $0.996$ \\ 
{\scriptsize SVHN and CIFAR100} & $0.773$ & $\mathbf{0.834}$ &$0.996$ & $0.996$ \\ 
{\scriptsize Tiny and SVHN} & $0.000$ & $\mathbf{0.996}$ &$0.219$ & $\mathbf{0.951}$ \\ 
{\scriptsize Tiny and CIFAR10} & $0.172$ & $\mathbf{0.708}$ &$0.882$ & $0.908$ \\ 
{\scriptsize Tiny and CelebA} & $0.012$ & $\mathbf{0.919}$ &$0.895$ & $0.895$ \\ 
{\scriptsize Tiny and CIFAR100} & $0.190$ & $\mathbf{0.701}$ &$0.880$ & $0.910$ \\ 
{\scriptsize CIFAR10 and SVHN} & $0.000$ & $\mathbf{0.987}$ &$0.064$ & $\mathbf{0.944}$ \\ 
{\scriptsize CIFAR10 and Tiny} & $0.065$ & $\mathbf{0.675}$ &$0.452$ & $0.458$ \\ 
{\scriptsize CIFAR10 and CelebA} & $0.000$ & $\mathbf{0.847}$ &$0.360$ & $\mathbf{0.648}$ \\ 
{\scriptsize CIFAR10 and CIFAR100} & $0.120$ & $\mathbf{0.688}$ &$0.528$ & $0.560$ \\ 
{\scriptsize CelebA and SVHN} & $0.001$ & $\mathbf{0.985}$ &$0.087$ & $\mathbf{0.747}$ \\ 
{\scriptsize CelebA and Tiny} & $0.130$ & $\mathbf{0.716}$ &$0.844$ & $0.845$ \\ 
{\scriptsize CelebA and CIFAR10} & $0.164$ & $\mathbf{0.702}$ &$0.877$ & $0.878$ \\ 
{\scriptsize CelebA and CIFAR100} & $0.176$ & $\mathbf{0.700}$ &$0.876$ & $0.878$ \\ 
{\scriptsize CIFAR100 and SVHN} & $0.000$ & $\mathbf{0.994}$ &$0.045$ & $\mathbf{0.945}$ \\ 
{\scriptsize CIFAR100 and Tiny} & $0.039$ & $\mathbf{0.766}$ &$0.416$ & $\mathbf{0.465}$ \\ 
{\scriptsize CIFAR100 and CIFAR10} & $0.052$ & $\mathbf{0.791}$ &$0.470$ & $0.504$ \\ 
{\scriptsize CIFAR100 and CelebA} & $0.000$ & $\mathbf{0.902}$ &$0.340$ & $\mathbf{0.663}$ \\ 
    \bottomrule
    \end{tabularx}
\label{tab:boosting_AUC_DM}
\end{table}

\begin{table}[h!]\captionsetup{font=footnotesize}
\centering
\footnotesize
\caption{AUC-ROC (higher is better) at A-gen vs.\ B task; due to the extensive computation time required for the DM baselines, the tasks are executed on subsamples of size 512. \textbf{Notation}: $^\ast$ tasks where likelihoods alone do not exhibit pathological behaviour, $\ddagger$ methods that employ external information or auxiliary models. For each task, we bold the best performing model.}
\begin{tabularx}{0.9\textwidth}{l*{8}{Y}}
\toprule
~~~Trained on & \multicolumn{2}{c}{MNIST}$^\ast$ & \multicolumn{2}{c}{FMNIST} & \multicolumn{2}{c}{CIFAR10} & \multicolumn{2}{c}{SVHN}$^\ast$ \\
\cmidrule(r){2-3} \cmidrule(lr){4-5} \cmidrule(lr){6-7} \cmidrule(l){8-9}
OOD Dataset& {\scriptsize FMNIST} & {\scriptsize Omniglot} & {\scriptsize MNIST} & {\scriptsize Omniglot} & {\scriptsize SVHN} & {\scriptsize CelebA} & {\scriptsize CIFAR10} & {\scriptsize CelebA} \\
\midrule
NF Likelihood & $0.007$ & $0.000$ & $0.000$ & $0.000$ & $0.014$ & $0.239$ & $\mathbf{0.970}$ & $\mathbf{0.984}$ \\
NF $\Vert\frac{\partial}{\partial \vect{x}} \log p_\theta(\vect{x}_0)\Vert_2$& $0.997$ & $0.997$ & $0.993$ & $0.993$ & $0.712$ & $0.379$ & $0.195$ & $0.077$ \\
Complexity Correction$^\ddagger$ & $0.026$ & $0.000$ & $0.044$ & $0.001$ & $0.678$ & $0.243$ & $0.714$ & $0.451$ \\ 
NF Likelihood Ratios$^\ddagger$ & $0.998$ & $\mathbf{1.000}$ & $\mathbf{1.000}$ & $\mathbf{1.000}$ & $0.299$ & $0.396$ & $0.302$ & $0.099$ \\
NF Dual Threshold (Ours)& $\mathbf{0.999}$ & $\mathbf{1.000}$ & $0.999$ & $0.996$ & $\mathbf{0.950}$ & $\mathbf{0.724}$ & $\mathbf{0.970}$ & $\mathbf{0.984}$ \\
\midrule
DM Likelihood & $0.843$ & $0.925$ & $0.000$ & $0.777$ & $0.000$ & $0.001$ & $0.806$ & $0.625$\\
DM Reconstruction &$\mathbf{1.000}$ & $\mathbf{1.000}$ & $0.994$ & $\mathbf{0.998}$ & $0.817$ & $0.610$ & $\mathbf{0.927}$& $\mathbf{0.930}$\\
DM Likelihood Ratios$^\ddagger$ & $0.985$ & $0.991$ & $\mathbf{0.971}$ & $\mathbf{0.998}$ & $0.959$ & $\mathbf{0.835}$ & $0.441$ & $0.474$\\
DM Dual Threshold (Ours)& $0.843$ & $0.931$ & $0.880$ & $0.938$ & $\mathbf{0.966}$ & $0.797$ & $0.806$ & $0.625$ \\
\bottomrule
\end{tabularx}
\label{tab:results-table-gen}
\end{table}

\subsection{Evaluation Metric Details} \label{appx:optimal_roc_curves}
\paragraph{Formal Definition of the ROC Curve for Dual Threshold Classifiers} When the ROC graph does not follow a curve-like structure -- as is the case with dual threshold classifiers -- \emph{optimal ROC curves} are used to generalize traditional ROC curves \citep{liu2022consistent}. The optimal ROC curve is obtained by first establishing a partial order on the ROC graph; a classifier is ``better'' than another if it has both a smaller FPR and a larger TPR.
Furthermore, the Pareto frontier of a partial order is the set of all maximal elements, and in turn, optimal ROC curves are defined by interpolating along the Pareto frontier of this partial order.
For illustration, \autoref{fig:ROG_ROC_AUC} shows the frontier alongside the optimal ROC curve as the upper boundary of the ROC graph using red lines.
In OOD detection, we consider a finite dataset of $N$ samples, each being assigned a likelihood and LID, constituting the sets $\{\log p_\theta(\vect{x}^{(n)})\}_{n=1}^N$ and $\{\widehat{\LID}_\theta(\vect{x}^{(n)})\}_{n=1}^N$, respectively.
These values can be seen as ``features'' and the labels associated with a datapoint are binary: whether the corresponding datapoint is in-distribution or OOD.
With that in mind,
the possible achievable combinations of FPR-TPR pairs are finite, making the Pareto frontier a set of disjoint points rather than a continuous curve. Therefore, we obtain the optimal ROC curve by step-interpolating the Pareto frontier of FPR-TPR pairs. The area under the optimal ROC curve holds various interpretations, making it an appropriate generalization of the traditional AUC-ROC in many model specification scenarios. For an in-depth exploration, we direct readers to \citet{liu2022consistent}.

\paragraph{Computing the Optimal ROC Curve} To numerically compute the optimal ROC curve, one must first define a set of dual threshold classifiers as:
\begin{equation}
    \Psi \coloneqq \{ (\psi_{\mathcal{L}} \pm \varepsilon, \psi_{\LID} \pm \varepsilon) : \psi_{\mathcal{L}} \in \{\log p_\theta(\vect{x}^{(n)}) \}_{n=1}^N , \psi_{\LID} \in \{\widehat{\LID}_\theta(\vect{x}^{(n)}) \}_{n=1}^N  \},
\end{equation}
where in practice we set $\varepsilon = 10^{-10}$. Since the cardinality of $\Psi$ is $\mathcal{O}(N^2)$, computing all FPR-TPR pairs may not be feasible. To address this, we select a subset of $\Psi$ and calculate the FPR-TPR pairs for this subset. We then compute their Pareto frontier, and in turn, the AUC-ROC. Although this method may underestimate the true AUC-ROC, we observe that the estimated AUC-ROC rapidly converges to the true value as the subset size increases. For the results presented in our tables, we have used a subset of $\Psi$ consisting of $5 \times 10^5$ classifiers.

\subsection{Extra Ablations} \label{appx:ablations}

Throughout our paper, we have argued in favour of our dual threshold method, which combines likelihoods and LID estimates. To highlight that our strong performance is not just based on dual thresholding itself, we carry out an ablation where we use dual thresholding, but on likelihood and gradient norm $\Vert \frac{\partial}{\partial \vect{x}} \log p_\theta(\vect{x}) \Vert_2$ pairs (or $\Vert s_\theta(\vect{x}, 0) \Vert_2$ for DMs). 
In this case, gradient norms are a proxy for how peaked a density is, in place of LID estimates. 
\autoref{tab:necessity_of_dual_thresholding_ablations} shows the results for both NFs and DMs, highlighting that LID estimates are much more useful. The table also shows that using single thresholds with LID estimates is also not enough to reliably detect OOD points. 
In the case of DMs, for the results presented in both \autoref{tab:necessity_of_dual_thresholding_ablations} and \autoref{tab:results-table}, we compute $s_\theta(\vect{x}, \epsilon)$ using a value of $\epsilon = 10^{-4}$. This approach is adopted to ensure numerical stability, a key consideration given that score-based diffusion models are known to become numerically unstable with extremely small timesteps \citep{pidstrigach2022score, lu2023mathematical}.

\begin{table}[t]
\centering
\caption{Ablation study on the necessity of dual threshold on AUC-ROC (higher is better).}
\label{tab:necessity_of_dual_thresholding_ablations}
\begin{tabularx}{0.9\textwidth}{p{5cm}|XXXX}
\toprule
{\footnotesize Method} & \leftcell{\scriptsize FMNIST vs.\ MNIST}& \leftcell{\scriptsize MNIST vs.\ FMNIST}& \leftcell{\scriptsize CIFAR10 vs.\ SVHN}& \leftcell{\scriptsize SVHN vs.\ CIFAR10}\\
\midrule
{NF \footnotesize $\LID_\theta(\vect{x})$ }& $0.951$ & $0.006$ & $0.936$ & $0.014$ \\
{\footnotesize NF $\Vert \frac{\partial}{\partial \vect{x}} \log p_\theta(\vect{x}) \Vert_2 \text{ and } \log p_\theta(\vect{x})$}& $0.516$ & $0.983$ & $0.722$ & $0.962$ \\
{\footnotesize NF $\LID_\theta(\vect{x}) \text{ and } \log p_\theta(\vect{x})$} & $0.951$ & $1.000$ & $0.936$ & $0.987$ \\
\midrule
{\footnotesize DM $\LID_\theta(\vect{x})$}& $0.912$ & $0.004$ & $0.944$ & $0.005$ \\
{\footnotesize DM $\Vert s_\theta(\vect{x}, 0) \Vert_2 \text{ and } \log p_\theta(\vect{x})$}& $0.240$ & $0.996$ & $0.883$ & $0.994$ \\
{\footnotesize DM $\LID_\theta(\vect{x}) \text{ and } \log p_\theta(\vect{x})$} & $0.912$ & $0.996$ & $0.944$ & $0.996$ \\
\bottomrule
\end{tabularx}
\end{table}

\subsection{Critical Analysis of OOD Baselines} \label{appx:baselines}
As we outlined in Section \ref{sec:experiments}, when benchmarking against the complexity correction and likelihood ratio methods, we observed notable underperformance in non-pathological directions. 
Both methods aim to correct inflated likelihoods encountered in pathological OOD scenarios by assigning a score to each datapoint, which is obtained by adding a complexity term to the likelihood \citep{serra2019input}, or subtracting a reference likelihood obtained from a model trained on augmented data \citep{ren2019likelihood}.
This score then becomes the foundation for their OOD detection through single thresholding. 
However, as we will demonstrate in this section, these techniques often necessitate an artificial hyperparameter setup to combine these metrics together, making it less than ideal. 

\begin{figure}[h!]\captionsetup{font=footnotesize}
    \centering
    
    \begin{subfigure}{0.24\textwidth}
        \includegraphics[width=\linewidth]{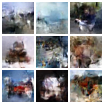} 
        \caption{Samples from RQ-NSF model trained on CIFAR10.}
    \end{subfigure}%
    \hfill
    \begin{subfigure}{0.24\textwidth}
        \includegraphics[width=\linewidth]{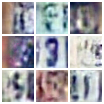}
        \caption{Samples from RQ-NSF model trained on SVHN.}
    \end{subfigure}%
    \hfill
    \begin{subfigure}{0.24\textwidth}
        \includegraphics[width=\linewidth]{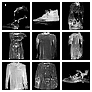}
        \caption{Samples from RQ-NSF model trained on FMNIST.}
    \end{subfigure}%
    \hfill
    \begin{subfigure}{0.24\textwidth}
        \includegraphics[width=\linewidth]{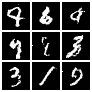}
        \caption{Samples from RQ-NSF model trained on MNIST.}
    \end{subfigure}

    \vspace{1em} 
    
    \begin{subfigure}{0.24\textwidth}
        \includegraphics[width=\linewidth]{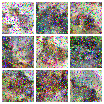}
        \caption{Samples from background model trained on CIFAR10.}
    \end{subfigure}%
    \hfill
    \begin{subfigure}{0.24\textwidth}
        \includegraphics[width=\linewidth]{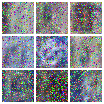}
        \caption{Samples from background model trained on SVHN.}
    \end{subfigure}%
    \hfill
    \begin{subfigure}{0.24\textwidth}
        \includegraphics[width=\linewidth]{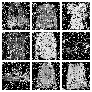}
        \caption{Samples from background model trained on FMNIST.}
    \end{subfigure}%
    \hfill
    \begin{subfigure}{0.24\textwidth}
        \includegraphics[width=\linewidth]{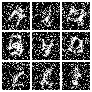}
        \caption{Samples from background model trained on MNIST.}
    \end{subfigure}
    
    \caption{Samples generated from normal and background models that are trained using the RQ-NSF hyperparameters provided in \autoref{tab:hyperparameters}. The background models are trained on perturbed data, using the scheme presented by \citet{ren2019likelihood}.}
    \label{fig:samples_rq_nsf_likelihood_ratio}
\end{figure}

Formally, both of these studies aim to find a score $\xi(\vect{x})$ to correct the inflated likelihood term $\log p_\theta(\vect{x})$, by adding a metric $m(\vect{x})$ as follows:
\begin{equation} \label{eq:likelihood-correction}
\xi(\vect{x}) = \log p_\theta (\vect{x}) + \lambda \cdot m(\vect{x}).
\end{equation}
In \citet{serra2019input}, $\lambda = 1$ and $m(\vect{x})$ is the bit count derived by compressing 
$\vect{x}$ using three distinct image compression algorithms and selecting the least bit count from the trio (an ensemble approach as they describe). 
The algorithms include standard \verb|cv2|, PNG, JPEG2000, and FLIF \citep{flif2021}. Moreover, we did not find any official implementation for the complexity correction method; however, since their algorithm was fairly straightforward, we re-implemented it according to their paper and it is readily reproducible in our experiments. On the other hand, \citet{ren2019likelihood} propose training a reference likelihood model with the same architecture as the original model; however, on perturbed data. We employ another RQ-NSF, samples of which are depicted in the bottom row of \autoref{fig:samples_rq_nsf_likelihood_ratio}. \citet{ren2019likelihood} claim that their reference model only learns background statistics that are unimportant to the semantics we care for in OOD detection; hence, subtracting the reference likelihood $m(\vect{x})$ can effectively correct for these confounding statistics that potentially inflate our original likelihoods. That said, they employ a hyperparameter tuning process on $\lambda$ to ensure best model performance.

\begin{figure}[h!]\captionsetup{font=footnotesize}
    \centering
    
    \begin{subfigure}{0.49\textwidth}
        \includegraphics[width=\linewidth]{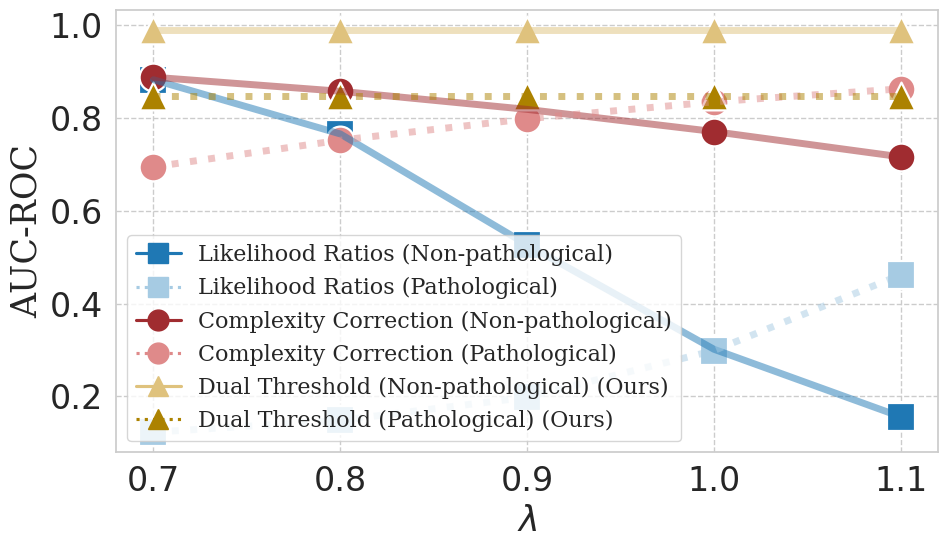} 
        \caption{Performance comparison of different methods on two pathological and non-pathological OOD detection tasks obtained from the FMNIST and MNIST pair.}
    \end{subfigure}%
    \hfill
    \begin{subfigure}{0.49\textwidth}
        \includegraphics[width=\linewidth]{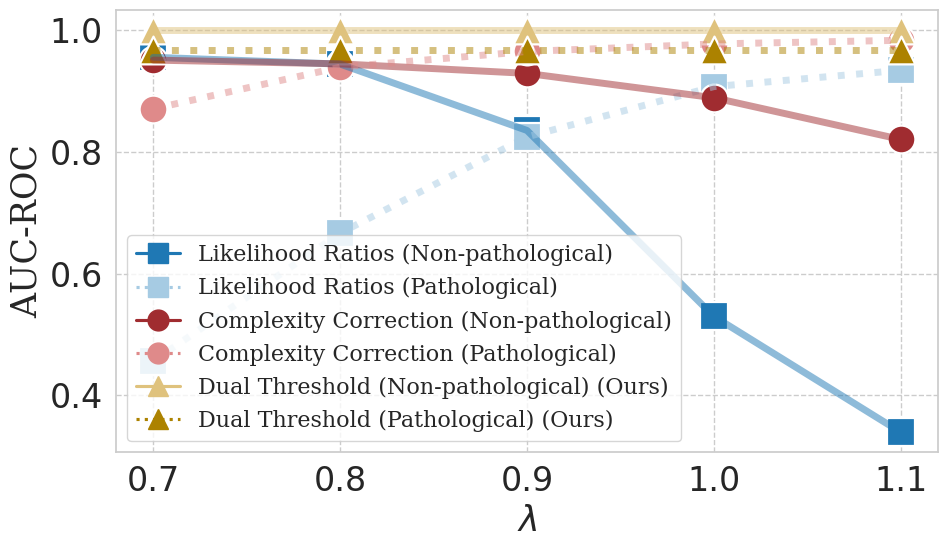}
        \caption{Performance comparison of different methods on two pathological and non-pathological OOD detection tasks obtained from the CIFAR10 and SVHN pair.}
    \end{subfigure}%
    
    \caption{
    Comparing our dual thresholding approach against all the different single score thresholding baselines by sweeping over different values of $\lambda$ in \autoref{eq:likelihood-correction}. The tasks that are considered are either: pathological such as \textbf{(a)} FMNIST vs.\ MNIST or \textbf{(b)} CIFAR10 vs.\ SVHN; or non-pathological such as \textbf{(a)}  MNIST vs.\ FMNIST or \textbf{(b)}  SVHN vs.\ CIFAR10.}
    \label{fig:sweep_lambda}
\end{figure}

As illustrated in \autoref{fig:sweep_lambda}, we sweep over values of $\lambda$ and compare our method against all these models. While certain $\lambda$ values enhance OOD detection in pathological scenarios, they falter in non-pathological contexts. In contrast, our dual thresholding remains robust irrespective of the scenario's nature. This observation underscores a significant gap in the OOD detection literature. While several methods address the OOD detection pathologies, many are overly specialized, performing well predominantly in the pathological direction. The results we report in \autoref{tab:results-table} correspond to the best values of $\lambda$.

\clearpage

\end{document}